\documentclass{article}

\usepackage[preprint]{neurips_2023}

\usepackage{multirow}
\usepackage{wrapfig}
\usepackage{subfigure}
\usepackage[utf8]{inputenc} 
\usepackage[T1]{fontenc}    
\usepackage{hyperref}       
\usepackage{url}            
\usepackage{booktabs}       
\usepackage{amsfonts}       
\usepackage{nicefrac}       
\usepackage{microtype}      
\usepackage{xcolor}         
\usepackage[pdftex]{graphicx}

\usepackage{amsmath}
\usepackage{amssymb}

\title{Enhancing Mapless Trajectory Prediction through Knowledge Distillation}

\author{~Yuning Wang{\small $~^{1}$}, ~Pu Zhang{\small $~^{2}$}, ~Lei Bai {\small $~^{3}$}, ~Jianru Xue{\small $~^{1}$}\\
\normalsize
$^{1}$\
Institute of Artificial Intelligence and Robotics, Xi'an Jiaotong University, China\\
\normalsize
$^{2}$\,  DiDi Chuxing, China\\
\normalsize
$^{3}$\, Shanghai AI Laboratory, China\\
\normalsize
\normalsize
{wangyn}@stu.xjtu.edu.cn,
\normalsize
\{zhangpu94,baisanshi\}@gmail.com,
\normalsize
jrxue@mails.xjtu.edu.cn
}

\begin{document}

\maketitle

\begin{abstract}
Scene information plays a crucial role in trajectory forecasting systems for autonomous driving by providing semantic clues and constraints on potential future paths of traffic agents. Prevalent trajectory prediction techniques often take high-definition maps (HD maps) as part of the inputs to provide scene knowledge. Although HD maps offer accurate road information, they may suffer from the high cost of annotation or restrictions of law that limits their widespread use. Therefore, those methods are still expected to generate reliable prediction results in mapless scenarios.
In this paper, we tackle the problem of improving the consistency of multi-modal prediction trajectories and the real road topology when map information is unavailable during the test phase. 
Specifically, we achieve this by training a map-based prediction teacher network on the annotated samples and transferring the knowledge to a student mapless prediction network using a two-fold knowledge distillation framework. Our solution is generalizable for common trajectory prediction networks and does not bring extra computation burden. Experimental results show that our method stably improves prediction performance in mapless mode on many widely used state-of-the-art trajectory prediction baselines, compensating for the gaps caused by the absence of HD maps. Qualitative visualization results demonstrate that our approach helps infer unseen map information.

\end{abstract}

\section{Introduction}
Trajectory prediction has been a critical task in autonomous driving scenarios [1], which bridges the upstream perception and downstream planning and control. Trajectory prediction aims at predicting the future movement of the concerned traffic participants according to the surrounding scene environment information and history trajectories of the traffic interactors. The accuracy of the trajectory prediction module largely influences the security of the autonomous driving system [2]. A number of advanced trajectory prediction methods have emerged in recent years [3,4,5,6].

High-definition maps (HD maps) have received wide attention as an important input in cutting-edge prediction methods. HD maps can provide detailed road topology and traffic semantics around the driving agent, 
which largely helps the predictor to understand scene information and traffic rules. 
The format of HD maps has transferred from rendered raster pixels with different semantic classes [7,32] to vectors [3], which provide more intensive information and need no rendering process.
Vectorized HD maps become widely used in many recent high-performance algorithms [3,4,5,6].
Also, large-scale datasets with rich HD maps have been constructed to better evaluate the map-based prediction methods [8,9]. Despite the changes in the specific format of HD maps, prevalent prediction methods mainly incorporate HD maps as part of the input to provide road knowledge.

The existing methods of utilizing maps may pose hidden dangers during test time.
While HD maps offer precise road information, their construction and semantic annotation within the traditional offline LiDAR-SLAM-annotation map construction pipeline require significant human effort [10]. Moreover, the maintenance cost for keeping HD maps up-to-date is considerable [28]. Consequently, concerns arise regarding the timeliness and coverage rate of HD maps, impeding their practical implementation in real-world test-time scenarios.
Furthermore, the qualification for HD map construction is highly restricted by laws in many countries or in certain critical areas [28].
Existing ways of using HD maps as inputs in trajectory prediction are not robust enough to address these issues.

To overcome the limitations of HD maps, some studies eliminate their dependence on offline annotated HD maps in trajectory prediction frameworks by exploiting online vectorized map construction using onboard sensors[10,11,12]. Despite their strong progress, online mapping methods are not specifically optimized for prediction, so they might spend efforts predicting unimportant maps, which is computationally cumbersome. Integrating online mapping into trajectory prediction frameworks presents challenges in terms of a large number of parameters and increased inference latency[12].
 Other studies perform end-to-end trajectory prediction without utilizing the annotated HD map in both the training and testing phase [13,14,15]. The end-to-end mapless prediction methods do not explicitly construct a map or predict a grid map and raster future trajectories, therefore might add computation burdens and lose the benefits of high definition and vectorization. And the jointly optimized framework makes it difficult to plug in the new advanced prediction algorithms. Furthermore, the end-to-end prediction pipeline wastes the existing annotated HD maps. Also, both of the above-mentioned two pipelines rely on perceptual data, which might be not reliable in extreme scenarios [15,30,33].

In this paper, we propose a new training framework for trajectory prediction methods which is different from present mapless prediction pipelines. Our framework leverages the benefits of training phase annotated HD maps while remaining robust to their flaws. We extract knowledge from training phase samples with annotated HD maps by training a map-based teacher network on them and generalize to the extreme test-time mapless situation by performing knowledge distillation (KD) to a mapless student network. Specifically, we design a two-fold knowledge distillation framework: feature distillation and output distillation (FOKD), which is based on the widely used encoder-decoder prediction network structure and inspired by the theories in [20] and [21].  The teacher features extracted from HD maps serve the subsequent prediction, and the teacher multi-hypothesis outputs must adhere to the map rules. As a result, both the encoded map features and the outputs contain certain knowledge that is beneficial for the student network to infer the underlying map. We exploit the knowledge from training time annotated HD maps by 
predicting a map feature and adding more supervision to the outputs.
Different from the end-to-end prediction methods, we take advantage of annotated HD maps and our framework is compatible with most prevalent trajectory networks, without building explicit grid maps.
Also different from online map construction which builds all HD map segments equally, our method can benefit from the knowledge aid from the teacher network and focus on the most critical part of HD maps, reducing the computation burden.
In our experiments, we use only agent history tracks as inputs and the same simple network structure as the teacher baseline prediction network, which is robust to the performance drop of sensors and can meet the efficiency demands of real-world scenarios.
Our proposed FOKD framework achieve stable improvements in test-time mapless prediction on various famous trajectory prediction baselines, demonstrating its effectiveness and generalization ability. 
Our framework itself brings nearly no additional computation burden or inference latency, and is also flexible enough to be enhanced with more input modalities or more complex network structures. 

\begin{figure*}
   \subfigure[End-to-end prediction]{
    \includegraphics[width=0.30\linewidth]{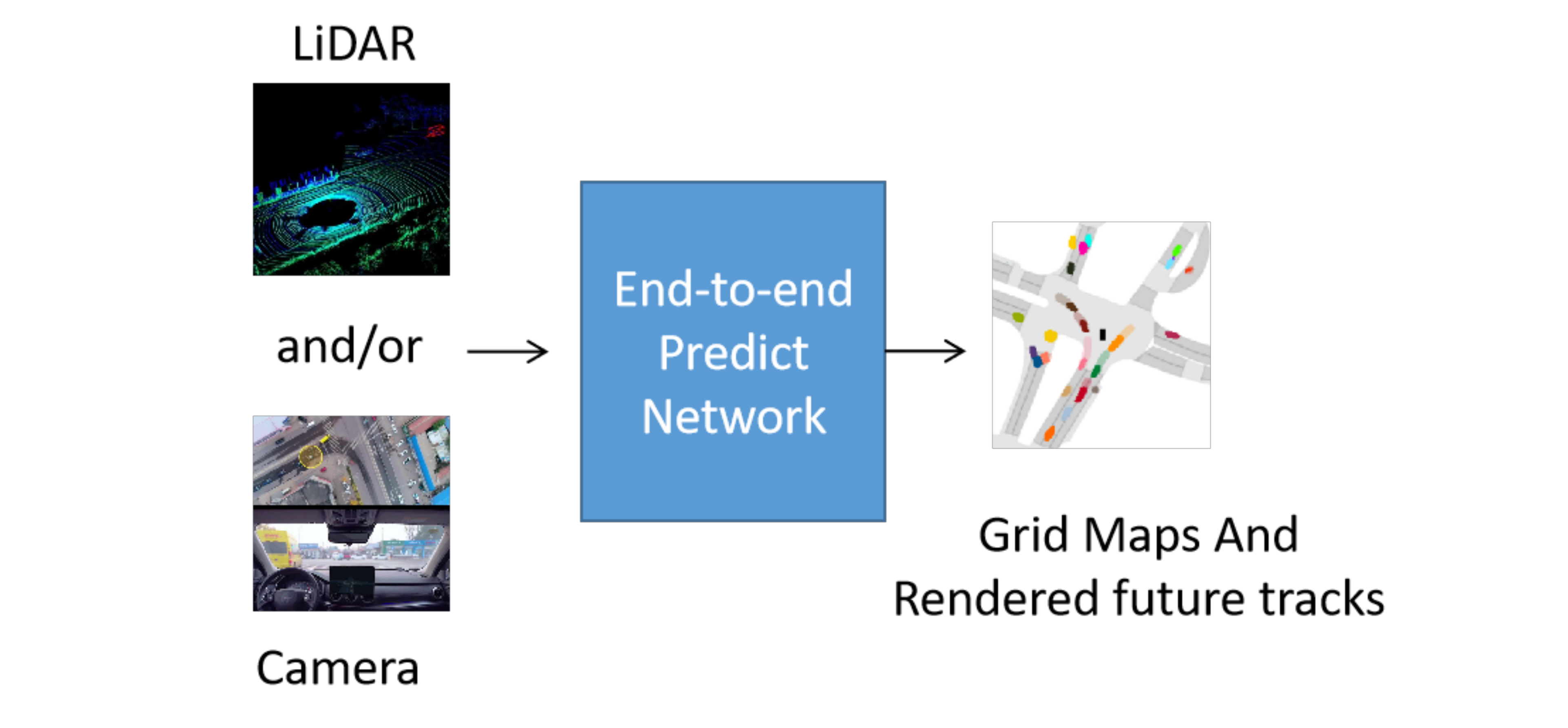} } 
    \subfigure[Online mapping]{
    \includegraphics[width=.30\linewidth]{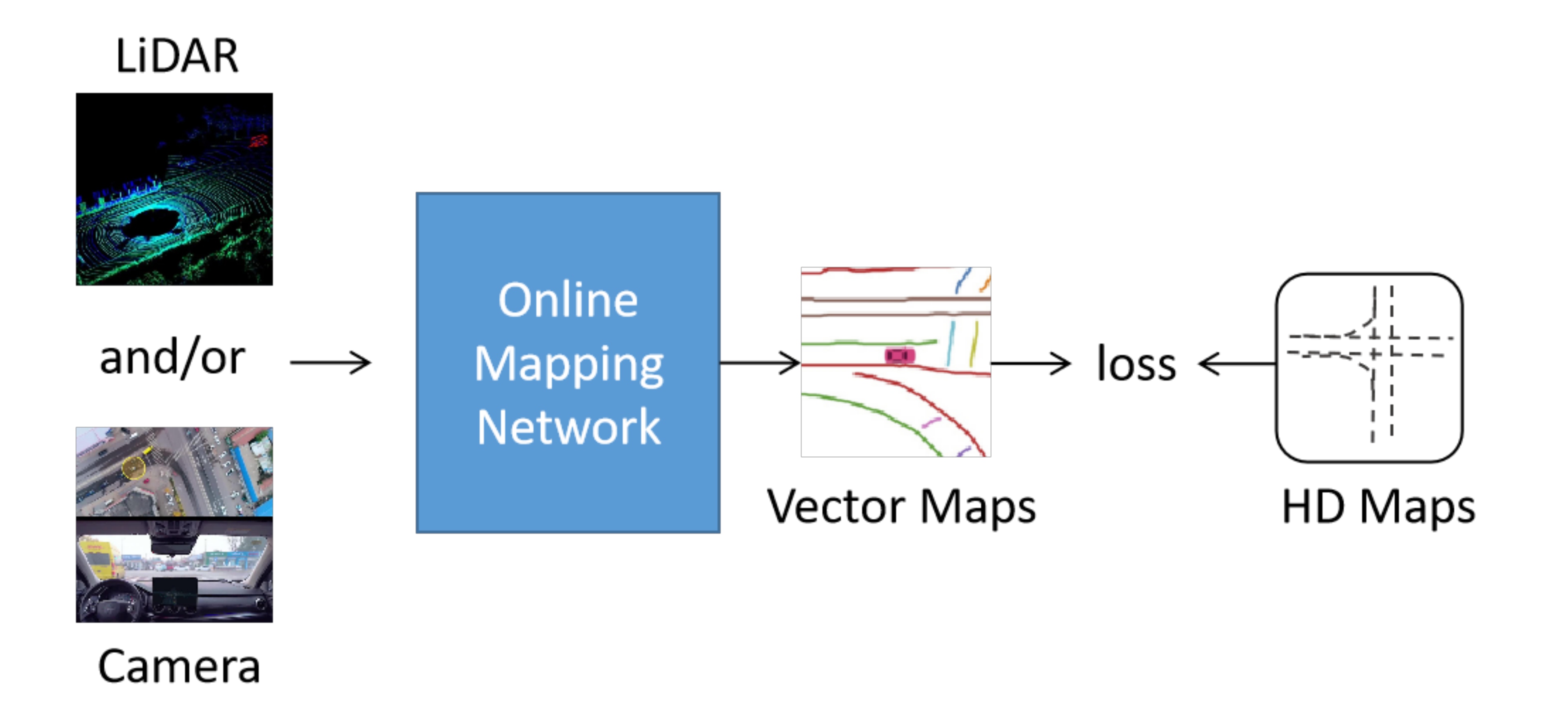} }
    \subfigure[Knowledge distillation (ours)]{
    \includegraphics[width=.30\linewidth]{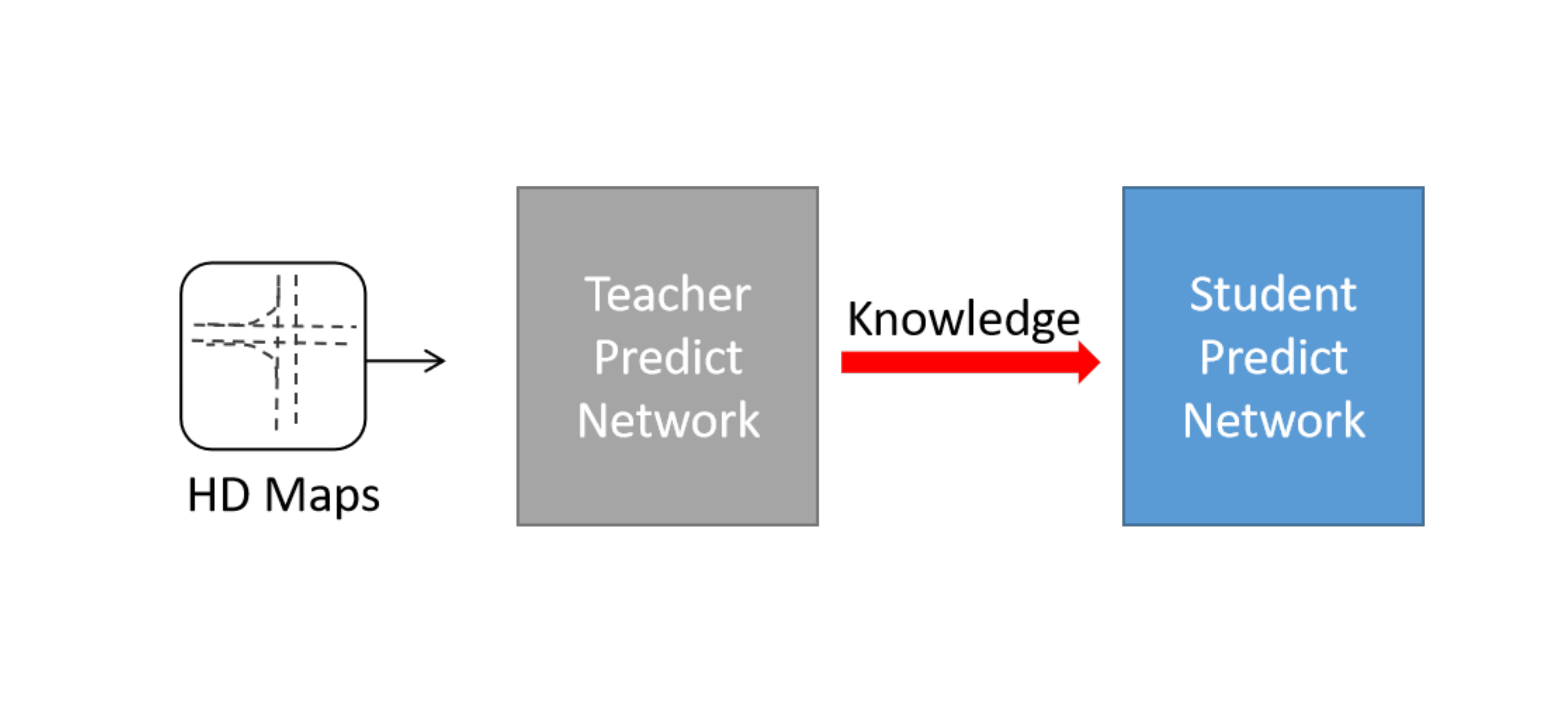} }
\caption{Two existing mapless prediction pipelines $\left(a\right)$, $\left(b\right)$ and our framework $\left(c\right)$. The three pipelines use annotated HD Maps in different ways: $\left(a\right)$ The end-to-end prediction networks do not use HD maps in both the training and testing phase. $\left(b\right)$ The online mapping methods use HD maps as labels to supervise the training phase. $\left(c\right)$ Our proposed knowledge distillation framework uses HD maps to training a map-based teacher predictor, and take the teacher knowledge as additional supervision in the training phase for a student mapless predictor.}    
\end{figure*}

To summarize, our contribution is three-fold:

(1) We propose a knowledge distillation framework for promoting the test-time mapless prediction performance. To our best knowledge, it is the first work to investigate how to exploit training phase HD maps for countering the test-time prediction performance degradation on mapless samples, which is different from existing mapless prediction pipelines.


(2) We design a two-fold KD method for the common encoder-decoder trajectory prediction network structures. We distill on both map features and network outputs, focusing on the critical map information filtered by the teacher. Our method can be applied to various trajectory prediction methods without difficulty.

(3) Experimental results show that our mapless prediction framework can stably make up for the test-time performance drop on various prediction benchmarks in the absence of HD maps, with nearly no additional computation burden and inference latency introduced.

\section{Related Works}

\textbf{Offline And Online HD Map Constructors} Traditional offline HD map is constructed based on the LiDAR point cloud outputs. The SLAM technique is adopted to fuse the LiDAR scans into a global consistent points cloud, then human annotation is required to build a global HD map with semantics.
This pipeline requires a lot of human effort, therefore limiting the coverage and freshness of HD maps. 
Being aware of the defects of offline HD maps, some works focus on constructing the local HD map around the agent online directly from the cameras and LiDAR
inputs, such as HDMapNet [10],  VectorMapNet [11], and MapTR [12]. 
Those methods have achieved high HD map construction accuracy, but at present directly incorporating them into prediction frameworks might introduce large inference latency and huge parameter amounts. 
To be more specific, the state-of-the-art online map generation will bring 40ms (MapTR[12])-100ms(VectorMapNet [11]) latency to the prediction network, which cannot meet the 20 ms inference latency limitation [6] in real scenarios. Also, present 2D-BEV modules in online mapping have around 30M parameters [12], which is relatively huge compare to aournd 3000K parameters in common prediction methods like HiVT [6]. 
In this work, our proposed specific mapless method brings nearly no additional parameters and latency. For stronger networks and more complex inputs, our FOKD framework can also implicitly choose to infer the underlying HD map that is critical to prediction with the aid from the teacher, which reduces the computation burden.  Our FOKD has the potential to be combined with online mapping.  

\textbf{Trajectory Prediction Using HD Maps} HD maps are extensively used in prevalent trajectory prediction methods, especially in vehicle trajectory prediction. At first HD maps are rendered into rasterized pixels with semantic meanings and encoded by a convolution network to be fed into the following prediction modules [7,32]. VectorNet [3] discovers the advantage of vectorized maps. The subsequent vector map based progress on predictions [4,5,6,16,17,18,19] further highlight the benefits of vectorization, producing strong baselines such as LaneGCN [4] and HiVT [6]. Some works proposed lane based decoder in a goal-based or regression form [16,17,18,19].
Also, many widely-used large-scale datasets like Argoverse [9] and Waymo [8], which have pre-annotate HD maps to be taken as input for prediction networks.
However, HD maps have some natural defects, such as huge annotation costs, hard to keep updated, and restriction of laws, which limit their use in real-world testing scenarios [28]. 
Without HD maps, those map-based decoders such as DenseTNT [18] are hard to apply, and the performance of the prediction framework with an HD map encoder will largely drop. In this work, we intend to transfer the knowledge of a map-based predictor to a mapless predictor to deal with the unavailability of HD map during the test time.

\textbf{Joint Trajectory Prediction And Map Generation} To circumvent the disadvantage of HD maps, Some works investigate end-to-end trajectory prediction as well as downstream planning [13,14,15]. 
Those works predict pixel-wise maps and future tracks though semantic segmentation or other techniques, which might not achieve high-definition and introduce computation burdens. Also the end-to-end pipelines are not compatible with advanced vectorized map based prediction solutions. Furthermore, the performance of the cameras or liDARs they rely on might be greatly affected in some extreme cases like dark, raining, or flashlights, doing harm to the end-to-end prediction performance [15,30]. 
In this work, our proposed detailed method maintains the network flexibility and can counter the performance drop on sensors,withour predicting grid map. The end-to-end methods do not use any HD maps in both the training and testing phase, therefore lose the advantage of HD maps: high-definition and vectorization, and can not take advantage of the exsiting annotated samples. Our FOKD framework is free of those flaws, and has potential to be combined with end-to-end pipelines.  

\textbf{Knowledge Distillation} Knowledge distillation (KD) is first brought up by Hinton \emph{etal.} [20] to transfer knowledge from a large-scale teacher network to a small student network to improve the performance of the small network. Hinton [20] distill on network outputs and FitNet [21] distills on mid-layer features. 
In this paper, we combine both the feature distillation and the output distillation to learn the latent map knowledge extracted by the teacher network.
Also, there has been some use of knowledge distillation in trajectory prediction for maintaining test time performance [25], taking the area history into account [26], and boosting scene-centric prediction [27]. 
None of them has discussed the unavailability of the HD maps, and we design a two-fold distilling method which is different from the literatures.

\section{Methods}

\textbf{Problem Formulation}
Trajectory prediction is a kind of
sequential prediction problems. Given observed past trajectories $X = \{(x^{i}_t,y^{i}_t), t=-T_{obs} +1,-T_{obs} +2,...,0\}^N_{i=1}$ of all the $N$ agents around a certain area and local HD map $M$, the objective is to predict the future coordinates of the target agent $\hat{Y}=\{(v, t=1,2, ..., T_{pred}\}$. In our setting, the test-time HD maps $M_{test}$ is not available , while the pre-annotated HD map $M_{train}$ can be utilized during training phase.


\begin{wrapfigure}{R}{0.5\linewidth}
\includegraphics[width=7.0cm]{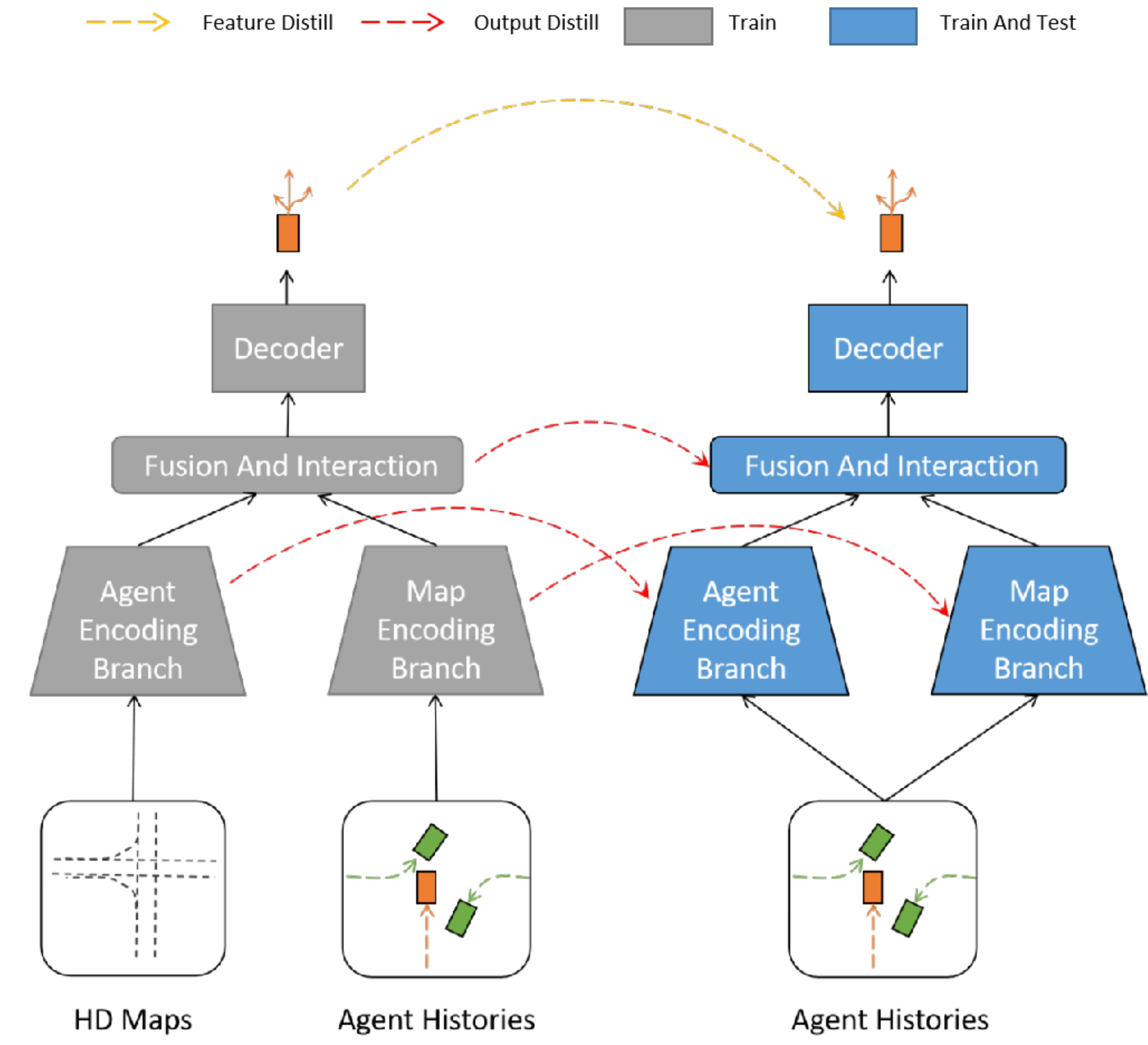}
\caption{Overview of our FOKD (Feature and Output Knowledge Distillation) framework. The network with gray modules represents the teacher and the blue one is the student.}
\label{fig: framework}
\end{wrapfigure}

\subsection{Overview}
We propose a two-fold knowledge distillation framework: feature distillation in \ref{sec:Feature Distillation}  and output distillation in
 \ref{sec:Output Distillation} according to the widely used encoder-decoder prediction network structure.
 The overview of FOKD is illustrated in \ref{fig: framework}. Our framework naturally suits prevalent trajectory prediction 
networks. A common trajectory prediction network generally contains an agent encoding branch, a map encoding branch, an agent-map features fusing and interaction module, and a decoder. 
We perform feature distillation on the
 map branch feature $f_m$, the agent branch feature $f_a$, and the feature after the fusion-and-interaction module $f_f$.  Output distillation is performed on network outputs after the decoder.
 All the blue modules are exist both for training and test phase, and the gray modules are only for training and will be removed at test time. 

In our experiment,  while our FOKD framework is not restricted to the network structure and inputs, we adopt the student with the same network structure as the teacher and replace the HD maps input with agent history tracks. We believe that the historical trajectories of other agents in the neighborhood area can give a lot of clues about the road topology and traffic semantics. For example, lots of braking tracks ahead might indicate traffic control, and many cross-running cars might mean the existence of intersections. Our design is robust to the performance drop of sensors and brings nearly no inference latency. A small variational design has also been added into the FOKD framework to deal with the student imitating noise caused by the missing input modality.




\subsection{Feature Distillation }
\label{sec:Feature Distillation}
We make the concerned student features $\{f^s_m,f^s_a,f^s_f\}$
all match the corresponding features in the teacher network $\{f^t_m,f^t_a,f^t_f\}$. 
In our setting, we are faced with missing input modality, which might bring large noise to the student feature of some samples, making them largely deviate from the teacher features. We utilize the variational-based knowledge distillation method to filter out those overly difficult samples. Variational knowledge distillation is firstly proposed in [23]. The feature matching noise is modeled as Gaussian noise, and the negative logistic likelihood of the teacher feature is taken as the feature distillation loss, which can be written as follows:
\begin{equation}
L_{FKD} = -log(P(f^t|f^s, \delta^s)) =  \frac{1}{2}*log({\delta^s}^2) + \frac{(f^t-f^s)^2}{{\delta^s}^2} + Z,
\label{eq:feature distill}
\end{equation}


where $f^t$ means an arbitrary teacher feature in the distillation set $\{f_m,f_a,f_f\}$  and $f^s$ means the corresponding student feature. $\delta^s$ is the variance of the Gaussian feature distribution, which is output by an additional branch only for the training phase and has the same shape as $f^s$. $Z$ is a normalization constant. We can see from Eq. \ref{eq:feature distill} that ${\delta^s}^2$ performs as a self-adaptive dimension-wise reweighting factor for the original L2 loss in FitNet [21], which is different for each sample therefore can help filter the hard ones. And the first term in Eq. \ref{eq:feature distill} restricts the scale of $\delta^s$.

\subsection{Output Distillation}
\label{sec:Output Distillation}
There are mainly two kinds of trajectory decoders [31]: regression-based [3,4,6] and goal-based [17,18,19].
We design output distillation scheme for both of the two kinds of decoders. 

\textbf{Regression-based decoder}
Regression-based decoders output $K$ hypothesizes in a deterministic [3,4] or a probabilistic [6] manner. The deterministic prediction network outputs estimated future coordinates $\hat{Y_{i}}$ as well as probabilities of each guess $\pi_i$, $\{\hat{Y_{i}}= F_{i}(X,M), \pi_i\}, i=1...K$. The probability prediction network outputs a multi-modal mixture of Gaussian or Laplacian distribution $P_i$ with $K$ mixture components and the probabilities of each component $\pi_i$, which can be write as $\{P_i\left(\hat{Y_i}| \mu_i(X),\sigma_i(X) \right), \pi_i\}, i=1...K$.



Traditional prediction networks take the winner-take-all training strategy. But apart from the best one closest to the ground truth, the other output guesses might contain certain learned knowledge about the map structure and the traffic rules, which is originally indicated in [20]. Therefore for output distillation, all the $K$ outputs is used. We make the student $i$th output hypothesis match the corresponding $i$th teacher hypothesis. Adding order to the outputs might break the symmetry of the loss function, but we find it effective in practice. For probabilistic prediction framework like HiVT, the formulation of distillation loss is as follows:
\begin{equation}
             L_{OKD} = -log(\sum_{i=1}^{K} P_{i}\left(s_{i}^t|\mu_{i}^{s},\sigma_{i}^{'s} \right)) + CE\left(\frac{\pi^t}{\tau},\frac{\pi^s}{\tau}\right),
\label{eq:od}
\end{equation}

where $s^t$ is samples from the teacher output distribution, $\mu_{i}^{s}$ is the prediction means in the student network, $CE$ is cross entropy for logits, and $\tau$ is the distilling temperature same as in [20]. For decerministic methods, $\mu^{'s} = \hat{Y}$. 
 An additional output branch is added for the adjusting variance $\sigma_{i}^{'s}$ during the training phase.
 The first negative logistic likelihood term in Eq.\ref{eq:od} will be as in Eq. \ref{eq:guassian variance reweighing} for Gaussian and in Eq. \ref{eq:guassian variance reweighing} for Laplacian distribution.
 

 \begin{equation}
 -log(\sum_{i=1}^{K} P_{i}\left(s_{i}^t|\mu_{i}^{s},\sigma_{i}^{'s} \right)) = \frac{1}{2}*log({\sigma_{i}^{'s}}^2) + \frac{(s_{i}^t-\mu_{i}^{s})^2}{2*{\sigma_{i}^{'s}}^2} + Z,
\label{eq:guassian variance reweighing}
\end{equation}

 \begin{equation}
 -log(\sum_{i=1}^{K} P_{i}\left(s_{i}^t|\mu_{i}^{s},\sigma_{i}^{'s} \right)) = log({\sigma_{i}^{'s}}) + \frac{|s_{i}^t-\mu_{i}^{s}|}{{\sigma_{i}^{'s}}} + Z.
\label{eq:lap  variance reweighing}
\end{equation}

We can see from Eq. \ref{eq:guassian variance reweighing} and Eq. \ref{eq:lap variance reweighing}. that the Gaussian distillation loss works as the L2 distance between the teacher and student outputs, and the Laplacian distillation loss works as the L1 distance. Also, the variance $\sigma_{i}^{'s}$ works like a loss self-adaptive sample-wise reweighting factor with scale restriction, balancing the distillation loss and prediction loss same as discussed in section \ref{sec:Feature Distillation}

\textbf{Goal-based decoder}
For prediction baseline with goal-based decoder such as DenseTNT [18], it is very laborious for them to sample a dense goal set around the agent without a map or a drivable area prior, especially for the vehicles that have a large velocity. Therefore we adopt its Gaussian regression-based variant for no-map prediction. We first render the final goal points of the student regression outputs into student goal heat maps $\overline{P^s}$ according to the output mean and variance, and then take the binary cross entropy in Y-net between the student goal heat maps and teacher goal heat $P^t$ maps as the output distillation loss:
\begin{equation}
L_{OKD} = BCE(\overline{P^s}(\mathbf{g}),P^t(\mathbf{g})),
\label{eq:output distill}
\end{equation}
where $\mathbf{g}$ means the selected goal set with the top $N$ probability predicted by the teacher network. $BCE$ means binary cross entropy and $P^t$ means the goal heat map output by the teacher.

\subsection{Training Loss}
Both the feature distillation loss $L_{FKD}$ and the output distillation loss $L_{OKD}$ act as auxiliary losses in the training phrase, and the training loss is as follows:

 \begin{equation}
L = L_{pred} + \lambda_{fd}*L_{FKD}+\lambda_{od}*L_{OKD},
\label{eq:lap  variance reweighing}
\end{equation}
where $L_{pred}$ is the prediction loss, and the $\lambda_{fd}$ and the $\lambda_{od}$ are rescaling factors for feature KD and output KD respectively.

\section{Experimental Results}

\subsection{Dataset}
We evaluate our proposed no-map prediction framework on a widely used large-scale vehicle dataset with rich HD maps: Argoverse [9]. Argoverse is a large-scale dataset collected in Pittsburgh and Miami with over 30K scenarios.
There are many state-of-the-art trajectory predicting works [3,4,5,6,17,18,19] which benchmark on Argoverse.  We take the common evaluation protocol of observing 2s and predicting 3s on Argoverse.

\subsection{Evaluation metrics}
We use the standard metrics for evaluating multi-modal trajectory prediction performance: Average-Displacement-Error (ADE), Final-Displacement-Error (FDE), and Miss Rate (MR), which are the official metrics of Argoverse. ADE means the averaged L2 distance between future prediction and ground truth trajectory, while FDE means the L2 distance between the final predicted destination and the ground truth destination. For evaluating multi-modality, we calculate minimum ADE and FDE
among all the output $K$ guesses, which are denoted as $ADE_{K}$ and $FDE_{K}$ and are averaged across the dataset. ADEs and FDEs are calculated in meters. Miss Rate refers to the ratio of test trajectories with FDEs of the best guess above 2.0 meters, and MR for $K$ guesses is denoted as $MR_{K}$.

\subsection{Baseline Methods}
In this work, we experiment on two widely known strong trajectory prediction baseline [4,18] and a state-of-the-art prediction method [6] that performs the best among all the open-sourced prediction methods on Argoverse, which shows the generality of our approach.

\textbf{DenseTNT and VectorNet}
DenseTNT [18] is a goal-based method and achieves strong performance, which builds on VectorNet [3] encoder and adopts a goal heat map decoder sampling a dense goal set.
When the map is not available, the dense goal set in DenseTNT decoder will be very computationally expensive and cannot be applied in practice. 
Therefore we choose a probabilistic regression-based VectorNet variant as the mapless student network for the DenseTNT teacher.


\textbf{LaneGCN}
LaneGCN [4] make progress by taking the lane connection structure into account. LaneGCN performs well on Argoverse and have been widely used as a powerful baseline.

\textbf{HiVT}
HiVT [6] is a strong baseline that adopts a hierarchical transformer structure with a local encoder and a global encoder. HiVT achieves a start-of-the-art result and is the best open-sourced method on Argoverse, and it also has the best no-map performance according to the ablation result in the paper [6]. Also, HiVT is computationally efficient with a small number of parameters and faster inference time. In this work, we both experiment on HiVT-64 and HiVT-128.

\subsection{Implement Details}
We take the same network structure and hyper-parameter as the baseline teacher network. The experiments are conducted on NVIDIA GeForce RTX 3090 GPUs. For HiVT, we select $\lambda_{fd}$ initially as $10$ and $\lambda_{od}$ as 1, and the $\lambda_{fd}$ decays to $0.1$ after 10 epochs, and $\lambda_{od}$ decay 10 times at the 10,20,40th epoch. 
For Vectornet,we select $\lambda_{fd}$ initially as $1$ and $\lambda_{od}$ as 50, and the $\lambda_{fd}$ and $\lambda_{od}$ all decays to $0.1$ after 10 epochs. For LaneGCN, we initially select $\lambda_{fd}$ as 10 and $\lambda_{od}$ as 1, and the they are decayed to $0.01$ and $0.001$ after 30 epochs. We attach a MLP projector after the student map feauture $f^s_m$ to match the teacher map feature $f^t_m$. At the training phase, an additional MLP is attached to the second-last layer features of the corresponding $f$ or $\mu$ output branch, outputing the variance $\delta$ or $\sigma$ with the same size as the MLP for $f$ or $\mu$ . $\tau$ is selected as $0.5$. For goal based teacher, the selected goal point number $N$ is set as 100. 

\subsection{Results And Comparisons}

\begin{table*}[]
\footnotesize
\resizebox{0.999\textwidth}{!}{
\centering
\begin{tabular}{lllllll}
\toprule
                       & $ADE_{6}$  & $FDE_{6}$  & $MR_{6}$   & $ADE_{20}$  & $FDE_{20}$  & $MR_{20}$   \\
\midrule
H-128 w/o m       & 0.73 & 1.15 & 0.13 & 0.64 & 0.76 & 0.06 \\
H-128 w/o m +FOKD & 0.71$\color{red}\downarrow2.7\%$& 1.11$\color{red}\downarrow3.5\%$ & 0.11$\color{red}\downarrow15.4\%$& 0.61$\color{red}\downarrow4.7\%$ & 0.70$\color{red}\downarrow7.9\%$ & 0.05$\color{red}\downarrow17\%$ \\
H-128 [6]              & 0.66 & 0.97 & 0.09 & 0.61 & 0.63 & 0.04  \\
\midrule
H-64 w/o m        & 0.77 & 1.25 & 0.14 & 0.68 & 0.85 &0.07\\ 
H-64 w/o m +FOKD  & 0.74$\color{red}\downarrow3.9\%$ & 1.20$\color{red}\downarrow4\%$ & 0.13 $\color{red}\downarrow7\%$  & 0.64 $\color{red}\downarrow5.9\%$ & 0.75 $\color{red}\downarrow11.8\%$ & 0.06 $\color{red}\downarrow14.2\%$\\
H-64 [6]                & 0.69 & 1.04 & 0.10 &0.64&0.70&0.05   \\
\midrule
V w/o m          & 0.93 & 1.70 & 0.24 & 0.73& 1.09 & 0.11                                     \\
V w/o m + FOKD  & 0.80$\color{red}\downarrow14\%$ & 1.33 $\color{red}\downarrow22\%$ & 0.15    $\color{red}\downarrow37.5\%$ &0.66 $\color{red}\downarrow9.5\%$ & 0.84 $\color{red}\downarrow23\%$ &0.07  $\color{red}\downarrow36\%$                                \\
V  [3]      & 0.80 & 1.32 & 0.15 & 0.65& 0.82&0.06   \\
D  [18]               & 0.73 &1.05 & 0.10 &/&/&/  \\
\midrule
L w/o m         & 0.79 & 1.29 & 0.15 &0.66 &0.84 & 0.07\\
L w/o m + FOKD  & 0.77$\color{red}\downarrow2.5\%$ & 1.22$\color{red}\downarrow5.4\%$& 0.13$\color{red}\downarrow13.3\%$ &0.64 $\color{red}\downarrow3\%$ & 0.76 $\color{red}\downarrow9.5\%$& 0.05 $\color{red}\downarrow28.6\%$\\
L  [4]               & 0.71 & 1.08 & 0.10 &0.61&0.70&0.04\\
\bottomrule

\end{tabular}
}
\caption{Prediction performance on Argoverse validation set. H means HiVT, V is VectorNet , D is DenseTNT, and L is LaneGCN. w/o m means without HD map inputs. We alter the VectorNet according to the official open source [34] to produce $K$ outputs and each output is Gaussian. }
\label{tab:val experimental result}
    \vspace{-0.2cm}
\end{table*}

\begin{table*}[]
\footnotesize
\centering
\begin{tabular}{llllllll}
\toprule
                        &$ADE_{6}$  & $FDE_{6}$  & $MR_{6}$ &$brier-minFDE_{6} $ \\
\midrule
HiVT-128 w/o map        &0.91  & 1.54 & 0.20 &2.21\\
HiVT-128 w/o map +FOKD   &0.88 $\color{red}\downarrow3.3\%$ & 1.47$\color{red}\downarrow4\%$ & 0.18$\color{red}\downarrow10\%$ & 2.15 $\color{red}\downarrow2.7\%$ \\
HiVT-128 [6]               &0.80  & 1.23 & 0.14  & 1.90\\
\midrule
HiVT-64 w/o map        & 0.95 & 1.64  & 0.22 & 2.32\\
HiVT-64 w/o map +FOKD   & 0.92 $\color{red}\downarrow 3.1 \%$ & 1.59 $\color{red}\downarrow 3 \%$ & 0.21 $\color{red}\downarrow4.5 \%$ &2.27 $\color{red}\downarrow2.2\%$\\
HiVT-64 [6]                & 0.83  & 1.31 & 0.15 &1.97  \\
\midrule
VectorNet w/o map         &  1.16 & 2.22 & 0.37& 2.91 \\
VectorNet w/o map +FOKD  & 0.99 $\color{red}\downarrow 14.6 \%$ & 1.74 $\color{red}\downarrow 21.6 \%$& 0.24 $\color{red}\downarrow 35.1 \%$ &2.43 $\color{red}\downarrow 16.5 \%$\\
VectorNet [3]                &0.93  &1.54& 0.20 &2.24  \\
\midrule
laneGCN w/o map       &0.98  & 1.71 &0.23 & 2.40\\
laneGCN w/o map +FOKD  & 0.96 $\color{red}\downarrow 2\%$ & 1.63 $\color{red}\downarrow 4.7\%$ & 0.22 
$\color{red}\downarrow 4.3\%$&2.32 $\color{red}\downarrow 3.3\%$ \\
laneGCN [4]              &0.87  & 1.37 & 0.16 & 2.06 \\
\bottomrule
\end{tabular}
\caption{Prediction performance on Argoverse test set. For the results of HiVT with map, we report the testing performance reproduced using the official code [35] and is in correspondence with [36].}
\label{tab:test experimental result}
    \vspace{-0.2cm}
\end{table*}

\textbf{Quantitative Result.}
 The quantitative experiment results are in Table \ref{tab:val experimental result} and \ref{tab:test experimental result} . We can see in Table \ref{tab:val experimental result} that our proposed FOKD framework can bring \textbf{consistent} promotion on the non-map prediction performance on various well-know baselines, which largely shows the generality of our method. Also, our method can make improvement on \textbf{all} the main metrics, which indicates the effectiveness of our method to make better prediction. Overall, our approach achieves a significant improvement. We narrow the gap between map-based trajectory prediction and mapless prediction by at least $20\%$ and at most nearly $100\%$.  On HiVT-128, which is the best open-sourced mapless prediction baseline on Argoverse, we achieve $3.5\%$, $7.9\%$ , $15.4\%$ and $17\%$ improvement on $FDE_{6}$, $FDE_{20}$, ,$MR_{6}$, $MR_{20}$ on the validation set, and $4\%$ and $10\%$ on the $FDE_{6}$ and $MR_{6}$ test set. 
We can also see from the table that as the baselines became stronger the improvements will be slightly dropped. But stronger baseline needs more parameters so there might be a trade-off. The test set results in Table \ref{tab:test experimental result} show coincident improvements  with the validation set results in Table \ref{tab:val experimental result}

\begin{figure*}[]
\small
\centering
\subfigure{\includegraphics[width=0.7\linewidth]{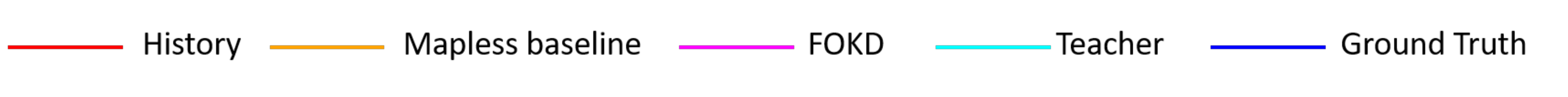}}
\subfigure{
\includegraphics[width=0.7\linewidth]{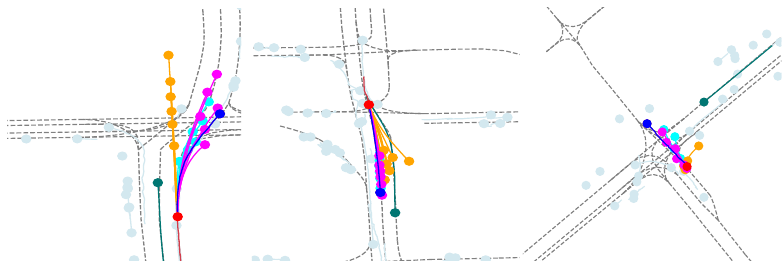}}
\caption{Qualitative visualizations on HiVT-128. We can see that compared to the mapless baseline our FOKD method can produce more map-aware results for turning, keeping lanes and waiting at intersections.}
\vspace{-0.2cm}
\label{fig:vis}
\end{figure*}

\textbf{Qualitative Results} Figure \ref{fig:vis} shows the qualitative comparisons with the mapless prediction baseline. We can see from the figure that our method can learn the map knowledge from the teacher network and implicitly infer the underlying HD map structure. Compared with the baseline mapless prediction network our method can produce prediction more compatible to the surrounding roads.

\begin{figure*}[t]
\small
\centering
    \includegraphics[width=0.7\linewidth]{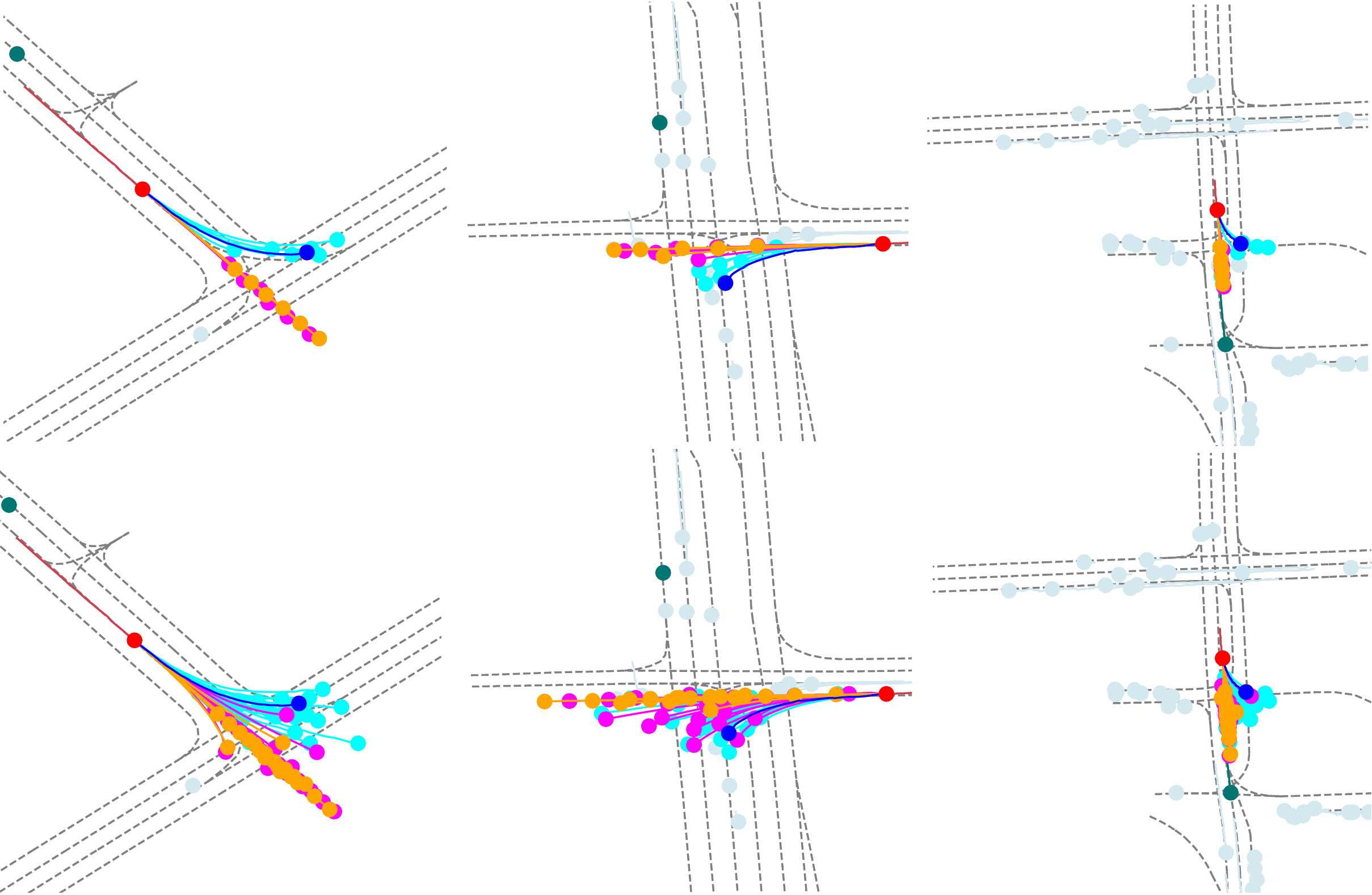}
\caption{Qualitative visualizations for 6 outputs (top row) vs 20 outputs (bottom row) on HiVT-128. We can see from the figure that adding more outputs will give the student more flexibility to catch the correct patterns, while baseline with out KD still shows no improvement}
\label{fig:vis_20}    
\end{figure*}

\begin{wrapfigure}{R}{0.5\linewidth}
\vspace{-0.8cm}
\includegraphics[width=7.0cm]{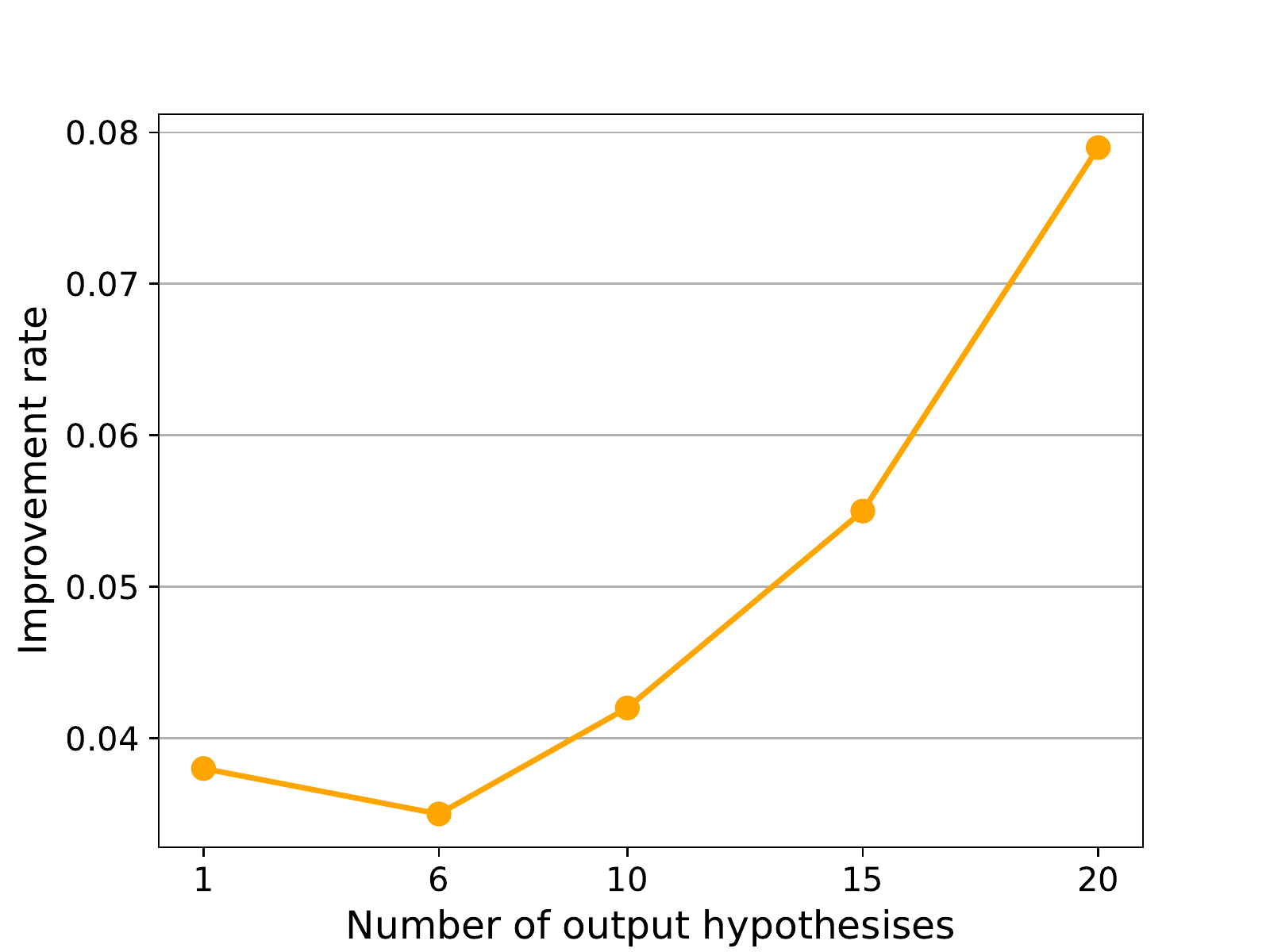}
\caption{Improvement rates on FDE for different $K$s on the mapless HiVT-128 student. We can see that the improvements grow along with the $K$s. }
\label{fig:poly}
\end{wrapfigure}

\textbf{Number of Outputs} We can see from Table \ref{tab:val experimental result} that the improvements on $K=20$ predictions is larger than the improvements on $K=6$ predictions. This is probably because that the mapless trajectory prediction  has more potential uncertainty.
HD map works as both clues and restrictions for possible future trajectories, and the inferred map features
may be not accurate and lack of details, therefore cannot eliminate the possibility of infeasible trajectories for some hard samples.
The $K=6$ metrics are for prediction with precise maps and cannot cover all those mapless uncertainty.  
For those hard samples, knowledge distillation will improve the probability of the correct motion pattern, but with less outputs $K$,  the wrong patterns with high probability still dominate.
Adding more outputs will give the decoder more flexibility so the correct pattern can be captured. Without the help of KD, the possibility of correct pattern is extreme low and still cannot be captured by larger $K$, as can been seen in the comparison between out method and the mapless baseline in Figure 
 \ref{fig:vis_20}. Also we can see the relationship between the $K$ and the improvement rate on $FDE$ for HiVT-128 in Figure 
 \ref{fig:poly}.

\subsection{Ablation And Parameter Sensitivity}
\begin{table*}[]
\footnotesize
\centering
\begin{tabular}{llllllll}
\toprule
Feature KD & Output KD & $ADE_{6}$  & $FDE_{6}$  & $MR_{6}$ & $ADE_{20}$ & $FDE_{20}$ & $MR_{20}$  \\
\midrule
           &           & 0.73 & 1.15 & 0.13 & 0.64 & 0.76 & 0.06 \\
\checkmark &           & 0.72 & 1.14 & 0.12 & 0.62 & 0.72 &  0.06  \\
           & \checkmark& 0.72 & 1.12 & 0.12 & 0.62&  0.71 &  0.05\\
\checkmark  &   \checkmark & \textbf{0.71} & \textbf{1.11} & \textbf{0.11} & \textbf{0.61}& \textbf{0.70}& \textbf{0.05}\\
\bottomrule
\end{tabular}
\caption{Ablations on the two proposed KD modules on HiVT-128 on Argoverse validation set}
\vspace{-0.3cm}
\label{tab:kd ablation}
\end{table*}

\begin{table*}[t]
    \begin{minipage}[t]{0.49\linewidth}
        \vspace{20pt}
    \centering
    \resizebox{0.999\textwidth}{!}{
\begin{tabular}{lllll}
\toprule
VR & $ADE_{6}$  & $FDE_{6}$  & $MR_{6}$ & $b-mFDE_{6}$   \\
\midrule
w/o                    & 0.88 & 1.48 & 0.19 & 2.18 \\
with           & \textbf{0.88} & \textbf{1.47} & \textbf{0.18} & \textbf{2.15} \\
\bottomrule
\end{tabular}}
\caption{Ablations on the variational reweighting modules on Argoverse test set on HiVT-128. VR represents variational reweighting and $b-mFDE_{6}$ represents $brier-minFDE_{6}$. }
\label{tab:va ablation}
    \end{minipage}
    \begin{minipage}[t]{0.49\linewidth}
    \vspace{0pt}
    \centering
    \resizebox{0.999\textwidth}{!}{
    \begin{tabular}{lllllll}
\toprule
$\lambda_{fd}$ & $\lambda_{kd}$& decay& $ADE_{6}$&$FDE_{6}$&$MR_{6}$\\
\midrule
0&0& N/A &0.73 & 1.15& 0.13\\
1&1& YES &0.71&1.12&0.12 \\
10&1& YES&\textbf{0.71}&\textbf{1.11}&\textbf{0.11}\\
1&10&YES&0.71&1.12&0.12\\
10&1&NO&0.73&1.16&0.12\\
1&10&NO&0.74&1.20&0.13\\
\bottomrule
\end{tabular}}
\caption{Parameter sensitivity studies on Argoverse validation set on mapless HiVT-128 student.}
\label{tab:param}
    \end{minipage}
    \vspace{-0.2cm}
\end{table*}

\textbf{Ablations} 
Table \ref{tab:kd ablation} shows ablations on our two proposed knowledge distillation modules. We can see from the table that the feature KD and the output KD can both achieve promotion alone, while using both of them can achieve better performance, showing the knowledge complementation of the teacher map features and outputs. Also, table \ref{tab:va ablation} shows effectiveness of our variation reweighting design.

\textbf{Parameter Sensitivity} 
Table \ref{tab:param} includes the parameter sensitivity study on the distillation loss factors $\lambda_{fd}$ and $\lambda_{od}$, as well as the study on the effect of afterwards decay of them during the training phase. It can be seen from the table that $\{10,1\}$ is the best choice, and different value set of $\lambda_{fd}$ and $\lambda_{od}$ can all achieve inprovements, indicating the robustness of our method. Also we can see that without the weight decay of $\lambda_{fd}$ and $\lambda_{od}$ the performances decline sharply. This might indicate that too strong distillations will harm the main prediction task in the later stages of training. Finding better multi-task balancing strategy might help.

\subsection{Limitations} In this paper we just use the trajectories around the target agent in the observed 2 seconds as the input. The limited information from these trajectories may impact the overall performance. Taking longer tracks or more modality as inputs and adopting stronger networks might bring further improvement but add more computation, and there should be a trade-off for realistic applications.

\section{Conclusion}
In this paper, we propose a two-fold knowledge distillation framework to address the test-time inexistence of HD maps for trajectory prediction, which is different from present pipelines.  Our experiments demonstrate a significant improvement in mapless performance using only agent history tracks, with nearly no additional inference latency. Our framework can also be augmented with stronger complex network structures, more input modalities, and longer observation, which is a compromise between effectiveness and efficiency.
\section*{References}
\medskip

{\small
[1] Houenou A, Bonnifait P, Cherfaoui V, et al. Vehicle trajectory prediction based on motion model and maneuver recognition[C]//2013 IEEE/RSJ international conference on intelligent robots and systems. IEEE, 2013: 4363-4369.

[2] Ren X, Yang T, Li L E, et al. Safety-aware motion prediction with unseen vehicles for autonomous driving[C]//Proceedings of the IEEE/CVF International Conference on Computer Vision. 2021: 15731-15740.

[3] Gao J, Sun C, Zhao H, et al. Vectornet: Encoding hd maps and agent dynamics from vectorized representation[C]//Proceedings of the IEEE/CVF Conference on Computer Vision and Pattern Recognition. 2020: 11525-11533.

[4] Liang M, Yang B, Hu R, et al. Learning lane graph representations for motion forecasting[C]//Computer Vision–ECCV 2020: 16th European Conference, Glasgow, UK, August 23–28, 2020, Proceedings, Part II 16. Springer International Publishing, 2020: 541-556.

[5] Ye M, Cao T, Chen Q. Tpcn: Temporal point cloud networks for motion forecasting[C]//Proceedings of the IEEE/CVF Conference on Computer Vision and Pattern Recognition. 2021: 11318-11327.

[6] Zhou Z, Ye L, Wang J, et al. Hivt: Hierarchical vector transformer for multi-agent motion prediction[C]//Proceedings of the IEEE/CVF Conference on Computer Vision and Pattern Recognition. 2022: 8823-8833.

[7] Hong J, Sapp B, Philbin J. Rules of the road: Predicting driving behavior with a convolutional model of semantic interactions[C]//Proceedings of the IEEE/CVF Conference on Computer Vision and Pattern Recognition. 2019: 8454-8462.

[8] Sun P, Kretzschmar H, Dotiwalla X, et al. Scalability in perception for autonomous driving: Waymo open dataset[C]//Proceedings of the IEEE/CVF conference on computer vision and pattern recognition. 2020: 2446-2454.

[9] Chang M F, Lambert J, Sangkloy P, et al. Argoverse: 3d tracking and forecasting with rich maps[C]//Proceedings of the IEEE/CVF conference on computer vision and pattern recognition. 2019: 8748-8757.

[10] Li Q, Wang Y, Wang Y, et al. Hdmapnet: An online hd map construction and evaluation framework[C]//2022 International Conference on Robotics and Automation (ICRA). IEEE, 2022: 4628-4634.
[11] Liu Y, Wang Y, Wang Y, et al. Vectormapnet: End-to-end vectorized hd map learning[J]. arXiv preprint arXiv:2206.08920, 2022.

[12] Liao B, Chen S, Wang X, et al. MapTR: Structured Modeling and Learning for Online Vectorized HD Map Construction[J]. arXiv preprint arXiv:2208.14437, 2022.

[13] Hu A, Murez Z, Mohan N, et al. FIERY: future instance prediction in bird's-eye view from surround monocular cameras[C]//Proceedings of the IEEE/CVF International Conference on Computer Vision. 2021: 15273-15282.

[14] Hu S, Chen L, Wu P, et al. St-p3: End-to-end vision-based autonomous driving via spatial-temporal feature learning[C]//Computer Vision–ECCV 2022: 17th European Conference, Tel Aviv, Israel, October 23–27, 2022, Proceedings, Part XXXVIII. Cham: Springer Nature Switzerland, 2022: 533-549.

[15] Kawasaki A, Seki A. Multimodal trajectory predictions for autonomous driving without a detailed prior map[C]//Proceedings of the IEEE/CVF Winter Conference on Applications of Computer Vision. 2021: 3723-3732.

[16] Narayanan S, Moslemi R, Pittaluga F, et al. Divide-and-conquer for lane-aware diverse trajectory prediction[C]//Proceedings of the IEEE/CVF Conference on Computer Vision and Pattern Recognition. 2021: 15799-15808.

[17] Zhao H, Gao J, Lan T, et al. Tnt: Target-driven trajectory prediction[C]//Conference on Robot Learning. PMLR, 2021: 895-904.

[18] Gu J, Sun C, Zhao H. Densetnt: End-to-end trajectory prediction from dense goal sets[C]//Proceedings of the IEEE/CVF International Conference on Computer Vision. 2021: 15303-15312.

[19] Gilles T, Sabatini S, Tsishkou D, et al. Home: Heatmap output for future motion estimation[C]//2021 IEEE International Intelligent Transportation Systems Conference (ITSC). IEEE, 2021: 500-507.

[20] Hinton G, Vinyals O, Dean J. Distilling the knowledge in a neural network[J]. arXiv preprint arXiv:1503.02531, 2015.

[21] Romero A, Ballas N, Kahou S E, et al. Fitnets: Hints for thin deep nets[J]. arXiv preprint arXiv:1412.6550, 2014.

[22] Tian Y, Krishnan D, Isola P. Contrastive representation distillation[J]. arXiv preprint arXiv:1910.10699, 2019.

[23] Ahn S, Hu S X, Damianou A, et al. Variational information distillation for knowledge transfer[C]//Proceedings of the IEEE/CVF Conference on Computer Vision and Pattern Recognition. 2019: 9163-9171.

[24] Zhao B, Cui Q, Song R, et al. Decoupled knowledge distillation[C]//Proceedings of the IEEE/CVF Conference on computer vision and pattern recognition. 2022: 11953-11962.

[25] Monti A, Porrello A, Calderara S, et al. How many observations are enough? knowledge distillation for trajectory forecasting[C]//Proceedings of the IEEE/CVF Conference on Computer Vision and Pattern Recognition. 2022: 6553-6562.

[26] Zhong Y, Ni Z, Chen S, et al. Aware of the History: Trajectory Forecasting with the Local Behavior Data[C]//Computer Vision–ECCV 2022: 17th European Conference, Tel Aviv, Israel, October 23–27, 2022, Proceedings, Part XXII. Cham: Springer Nature Switzerland, 2022: 393-409.

[27] Su D J A, Douillard B, Al-Rfou R, et al. Narrowing the coordinate-frame gap in behavior prediction models: Distillation for efficient and accurate scene-centric motion forecasting[C]//2022 International Conference on Robotics and Automation (ICRA). IEEE, 2022: 653-659.

[28] Bao Z, Hossain S, Lang H, et al. A review of high-definition map creation methods for autonomous driving[J]. Engineering Applications of Artificial Intelligence, 2023, 122: 106125.

[29] Behley J, Garbade M, Milioto A, et al. Semantickitti: A dataset for semantic scene understanding of lidar sequences[C]//Proceedings of the IEEE/CVF international conference on computer vision. 2019: 9297-9307.

[30] Zang S, Ding M, Smith D, et al. The impact of adverse weather conditions on autonomous vehicles: how rain, snow, fog, and hail affect the performance of a self-driving car[J]. IEEE vehicular technology magazine, 2019, 14(2): 103-111.

[31] Jiang B, Chen S, Wang X, et al. Perceive, Interact, Predict: Learning Dynamic and Static Clues for End-to-End Motion Prediction[J]. arXiv preprint arXiv:2212.02181, 2022.

[32]
Salzmann T, Ivanovic B, Chakravarty P, et al. Trajectron++: Dynamically-feasible trajectory forecasting with heterogeneous data[C]//Computer Vision–ECCV 2020: 16th European Conference, Glasgow, UK, August 23–28, 2020, Proceedings, Part XVIII 16. Springer International Publishing, 2020: 683-700.

[33] Ding W, Zhao J, Chu Y, et al. FlowMap: Path Generation for Automated Vehicles in Open Space Using Traffic Flow[J]. arXiv preprint arXiv:2305.01622, 2023.

[34] https://github.com/Tsinghua-MARS-Lab/DenseTNT

[35] https://github.com/ZikangZhou/HiVT

[36] https://github.com/ZikangZhou/HiVT/issues/14

\clearpage
\appendix

\section{Relationship Between Agent History Length And Knowledge Distillation Performance}
Figure \ref{fig:poly_traj} shows the relationship between the agent history lengths inputted into the map encoding branch and the knowledge distillation improvement in $FDE_{6}$ on HiVT-128. As the figure demonstrates, longer history lengths result in larger performance improvement, highlighting the significance of agent history information.

\begin{wrapfigure}{R}{0.5\linewidth}
\vspace{-0.8cm}
\includegraphics[width=7.0cm]{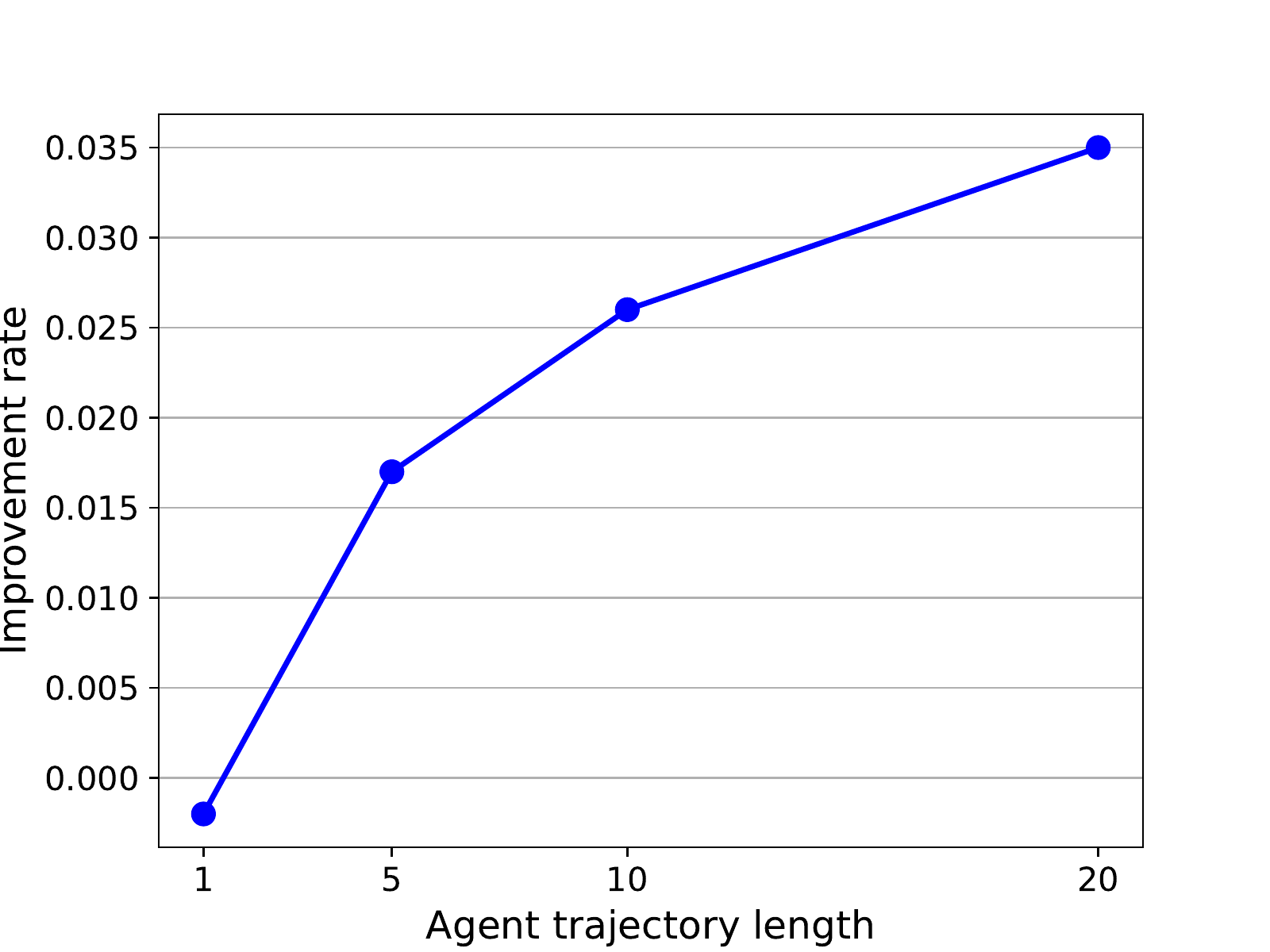}
\caption{Relationship between the agent history length and the $FDE_{6}$ improvement rate.}
\label{fig:poly_traj}
\end{wrapfigure}

\section{More Visualization}
More visualization of the comparison between our FOKD model and the mapless baseline model on HiVT-128 is in  Figure \ref{fig:vis_more}. Figure \ref{fig:vis_more} illustrates many scenarios where map information plays a critical role in future predictions, such as starting moving after waiting at the intersection (left column), turning at an intersection (middle column), and driving along the lane (right column). In those scenarios, our knowledge distillation method can better infer the knowledge of the underlying road structure information and therefore make better predictions than the mapless baseline. Also, we can see in figure \ref{fig:vis_more_20} for more visual comparison between 6 outputs and 20 outputs, which indicates that more outputs can lead to more improvement for our FOKD method, and the improvement of the baseline is small without knowledge distillation.
 
\begin{figure*}[h]
\centering
\subfigure{ \includegraphics[width=0.8\linewidth]{legend_nips2.pdf}}
   \subfigure{
    \includegraphics[width=0.30\linewidth]{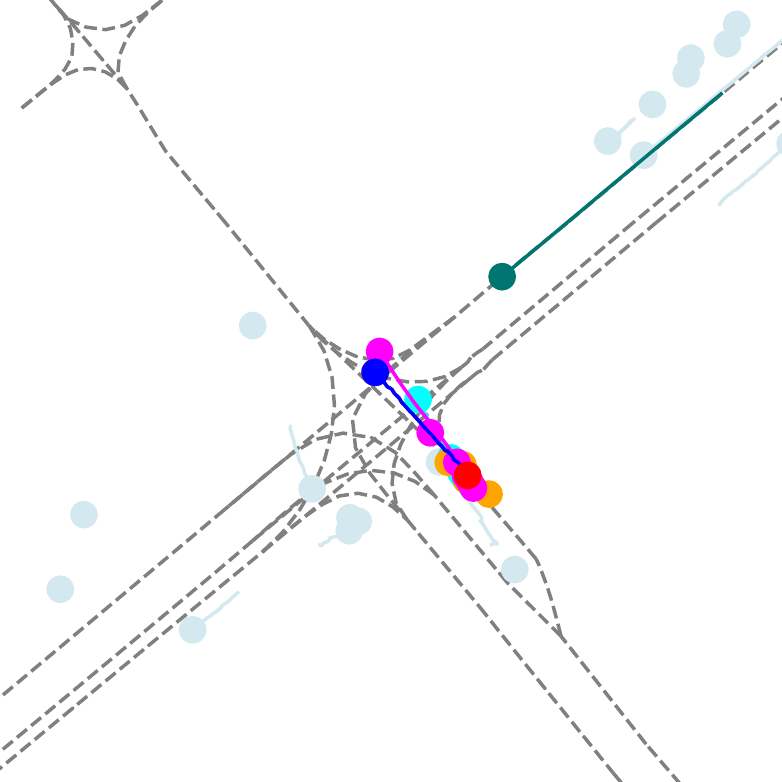} } 
    \subfigure{
    \includegraphics[width=.30\linewidth]{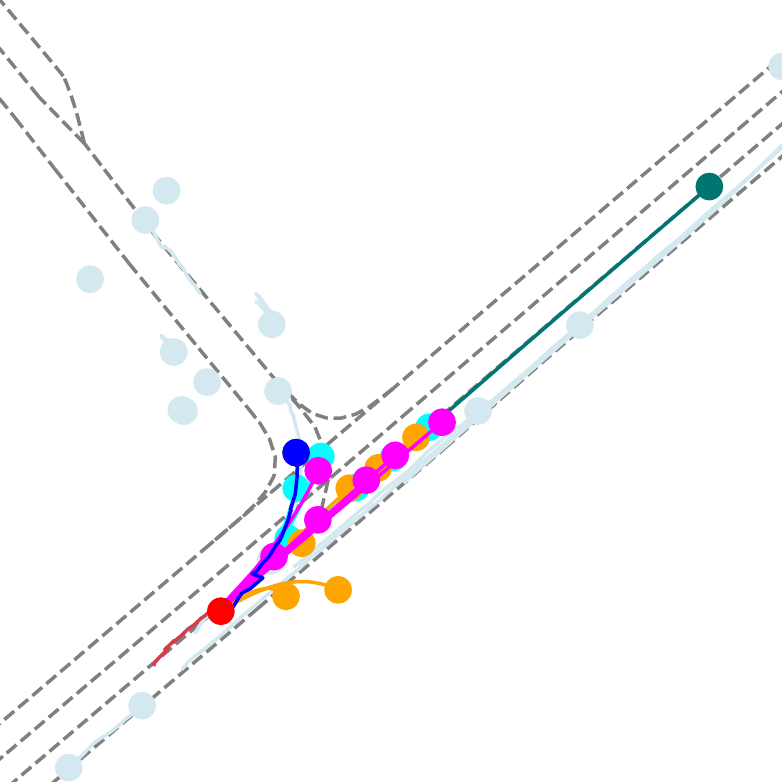} }
    \subfigure{
    \includegraphics[width=.30\linewidth]{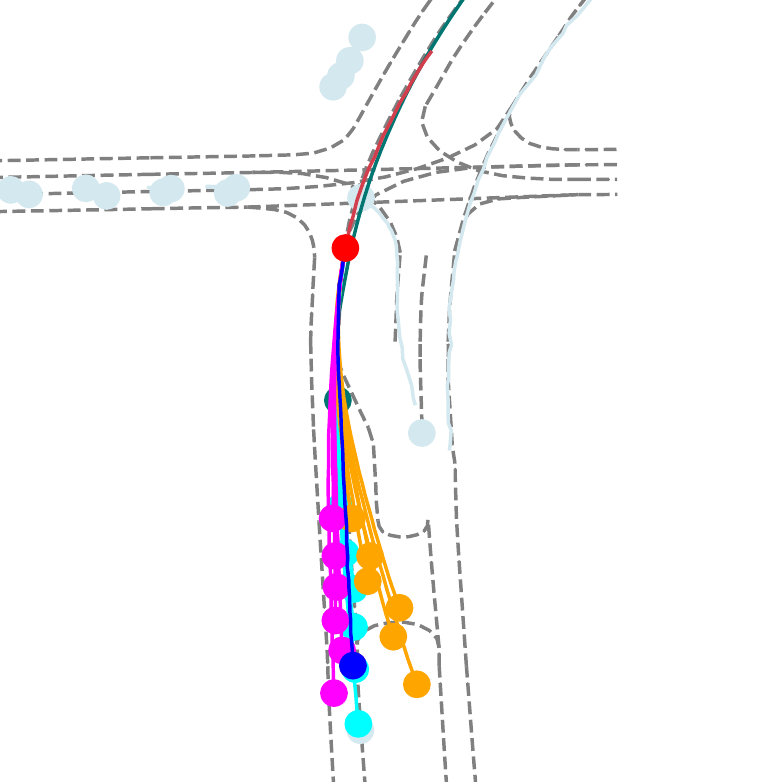} }
    \subfigure{
        \includegraphics[width=.30\linewidth]{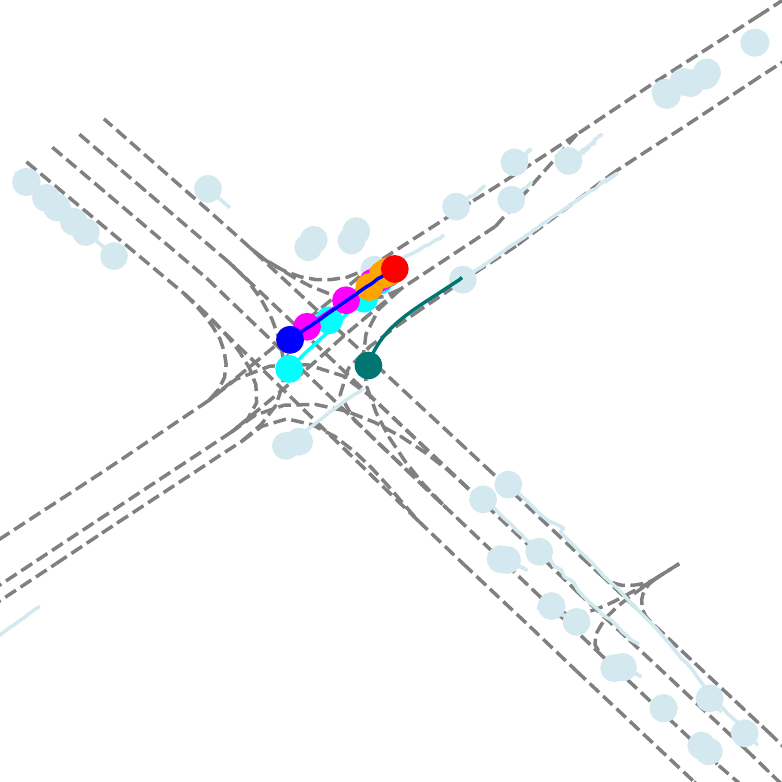} }
    \subfigure{
        \includegraphics[width=.30\linewidth]{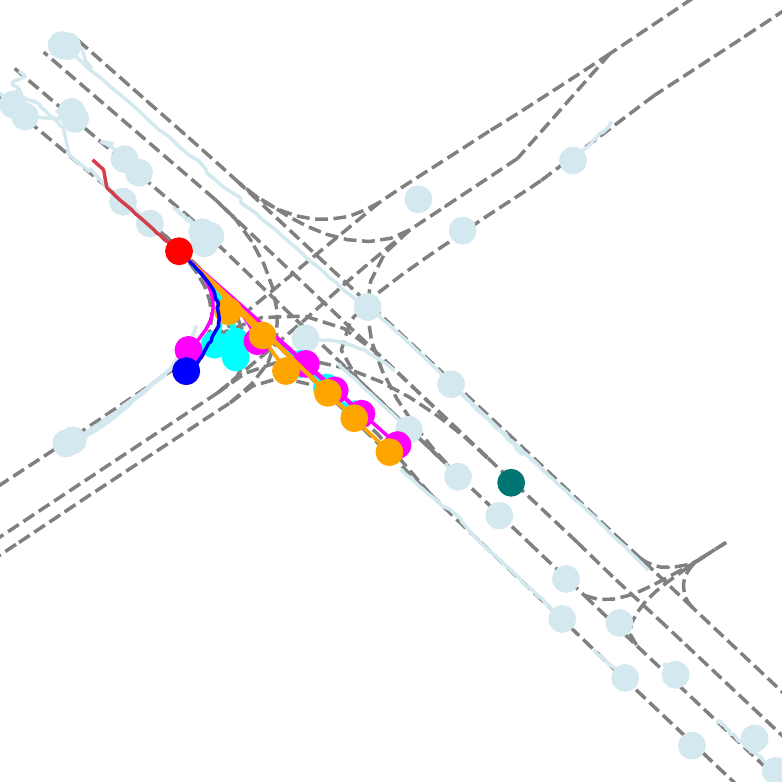} }
     \subfigure{
        \includegraphics[width=.30\linewidth]{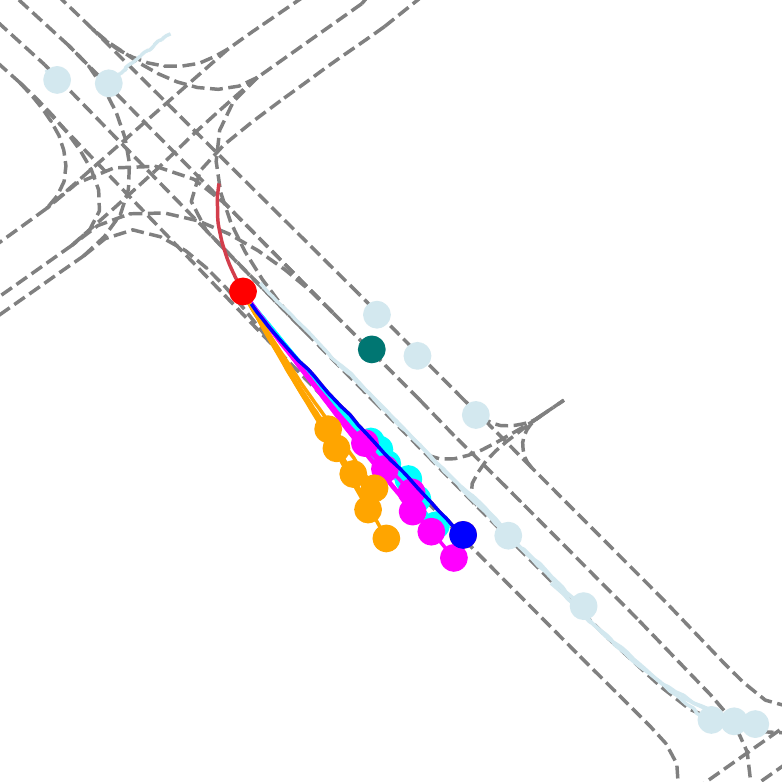} }     
    \subfigure{
        \includegraphics[width=.30\linewidth]{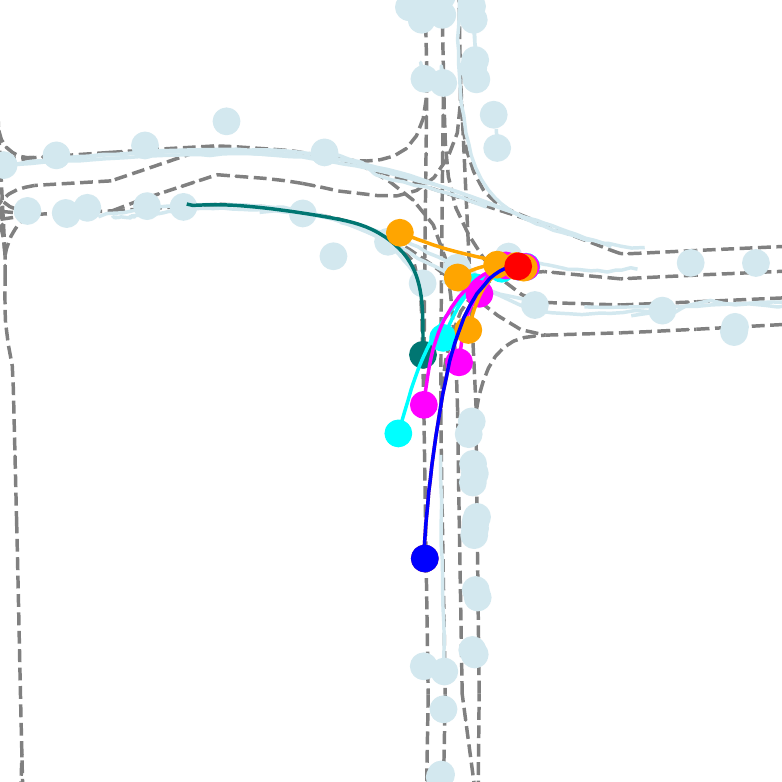} }  
        \subfigure{
        \includegraphics[width=.30\linewidth]{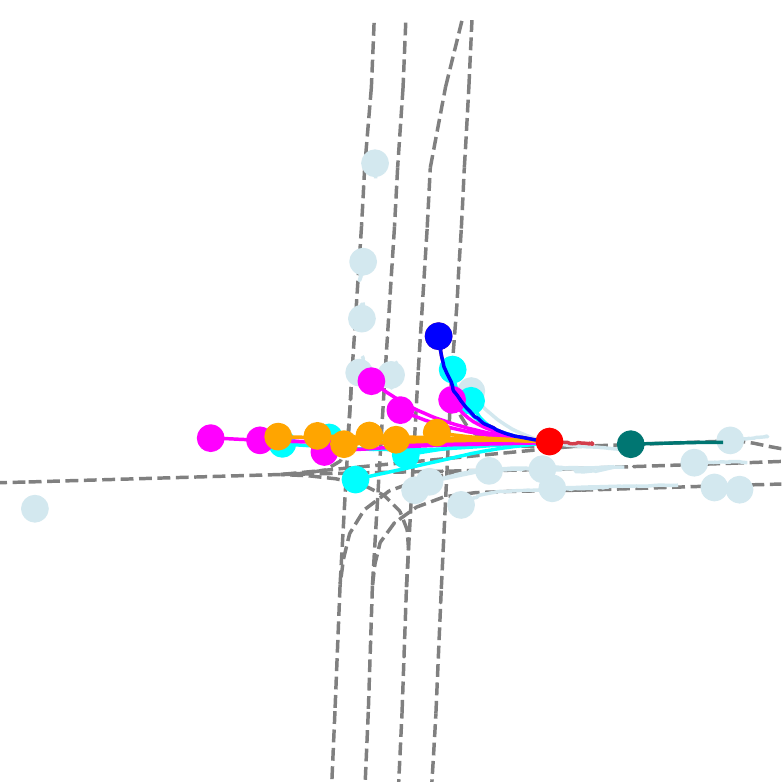} }     
            \subfigure{
        \includegraphics[width=.30\linewidth]{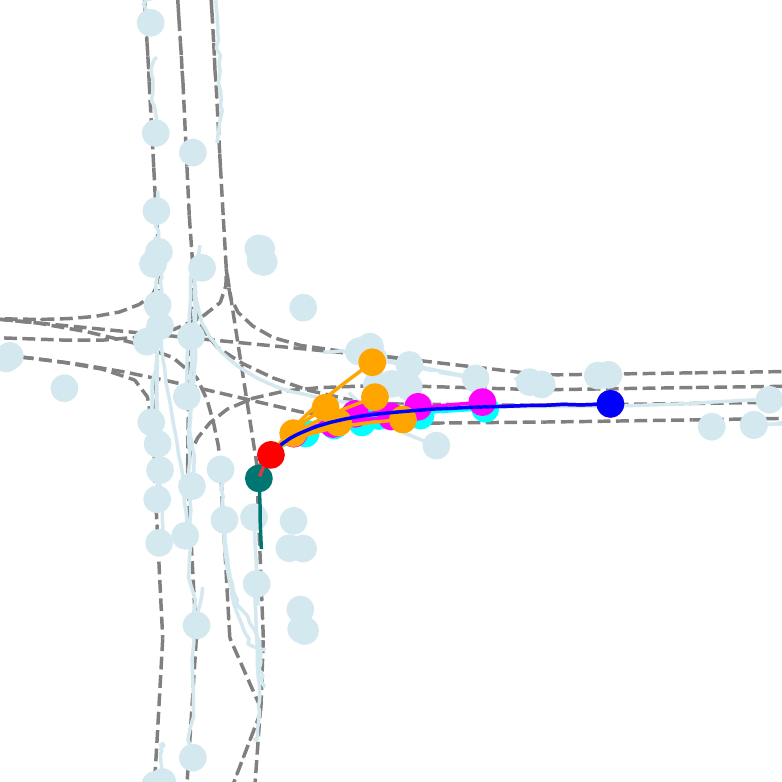} }   
\caption{More visualization results on HiVT-128.}  
\label{fig:vis_more}
\end{figure*}

\begin{figure*}
\centering
\subfigure{\includegraphics[width=0.30\linewidth]{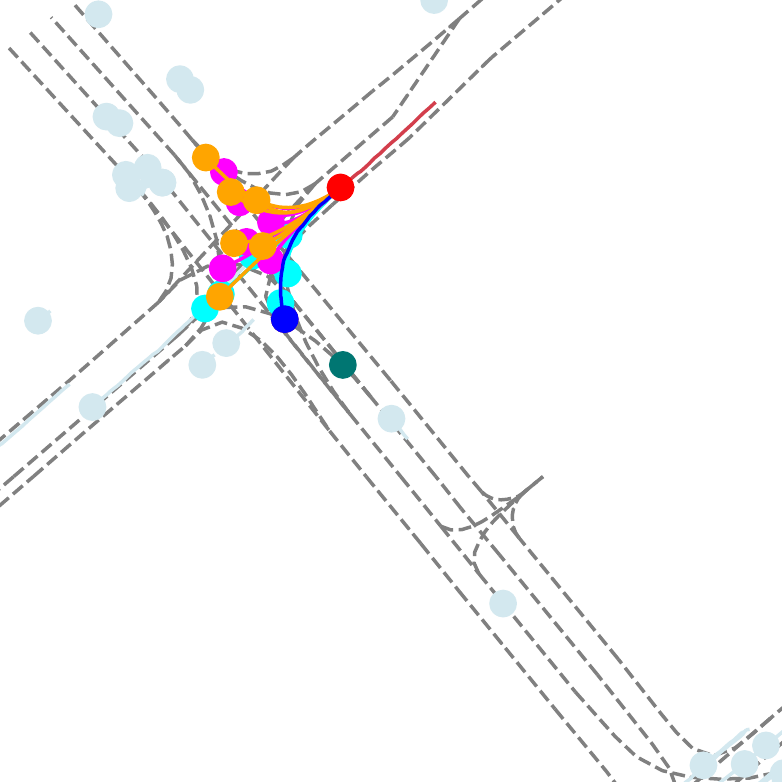} } 
\subfigure{\includegraphics[width=0.30\linewidth]{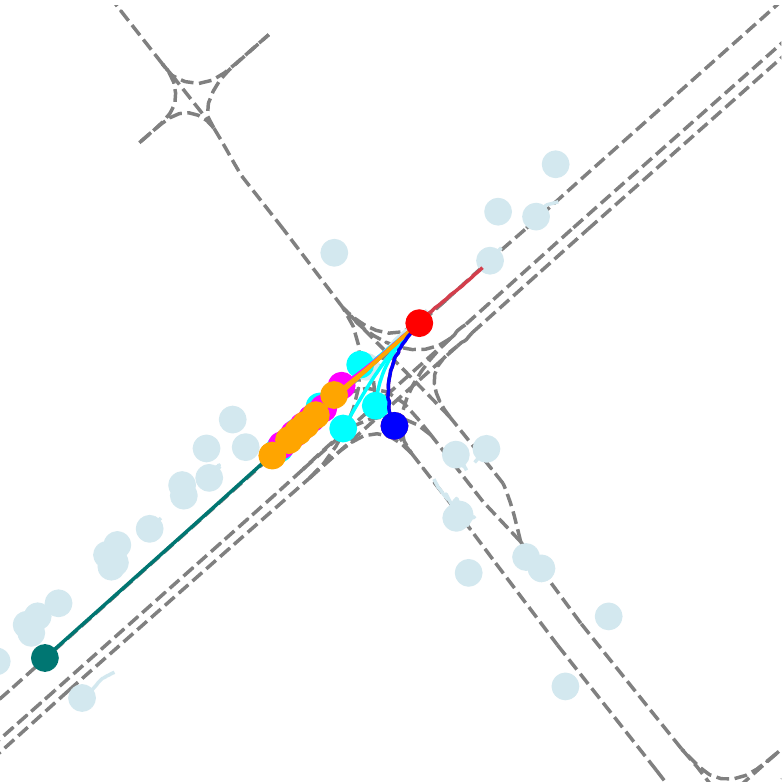} } 
\subfigure{\includegraphics[width=0.30\linewidth]{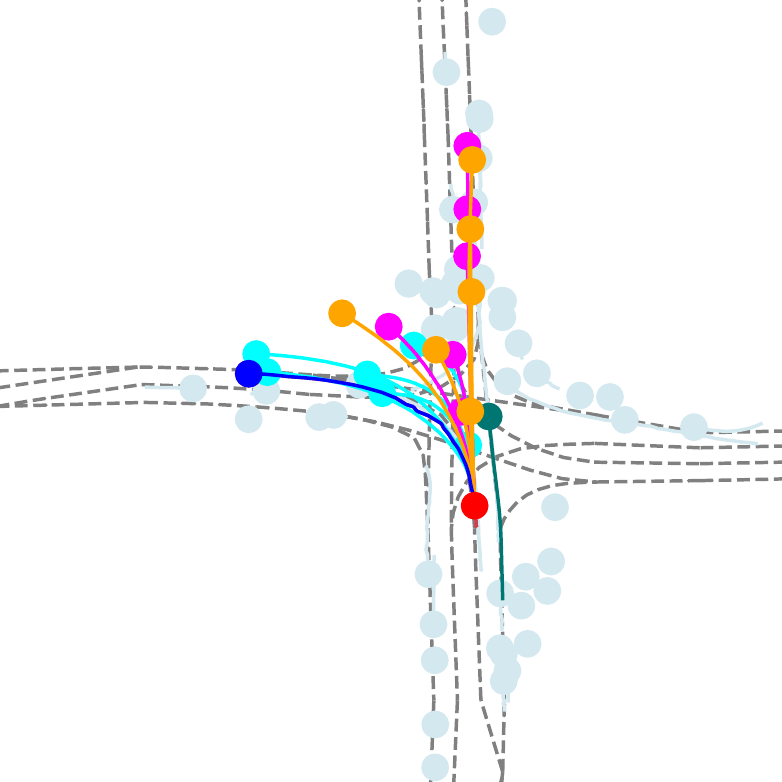} } 
\subfigure{\includegraphics[width=0.30\linewidth]{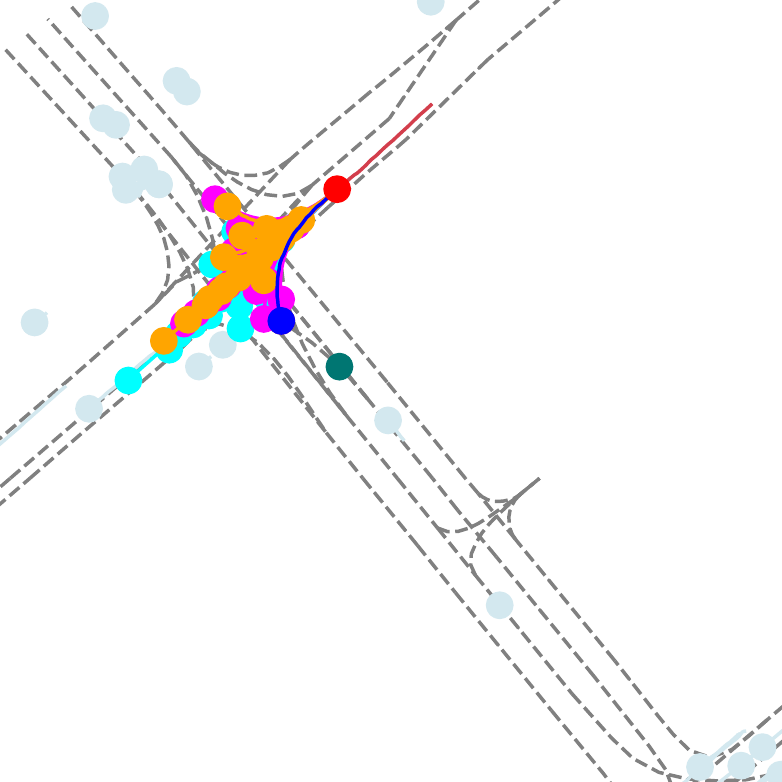} } 
\subfigure{\includegraphics[width=0.30\linewidth]{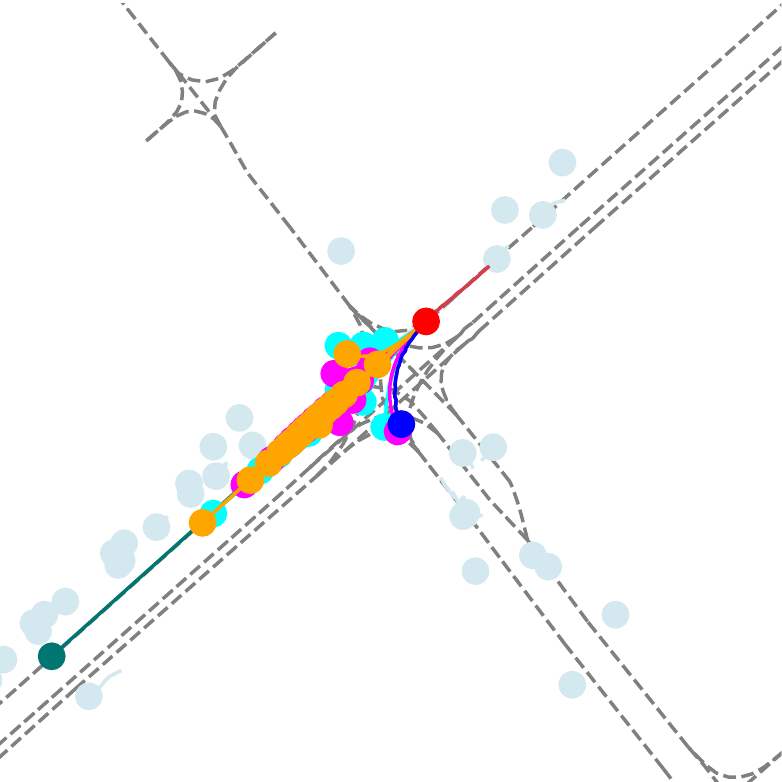} } 
\subfigure{\includegraphics[width=0.30\linewidth]{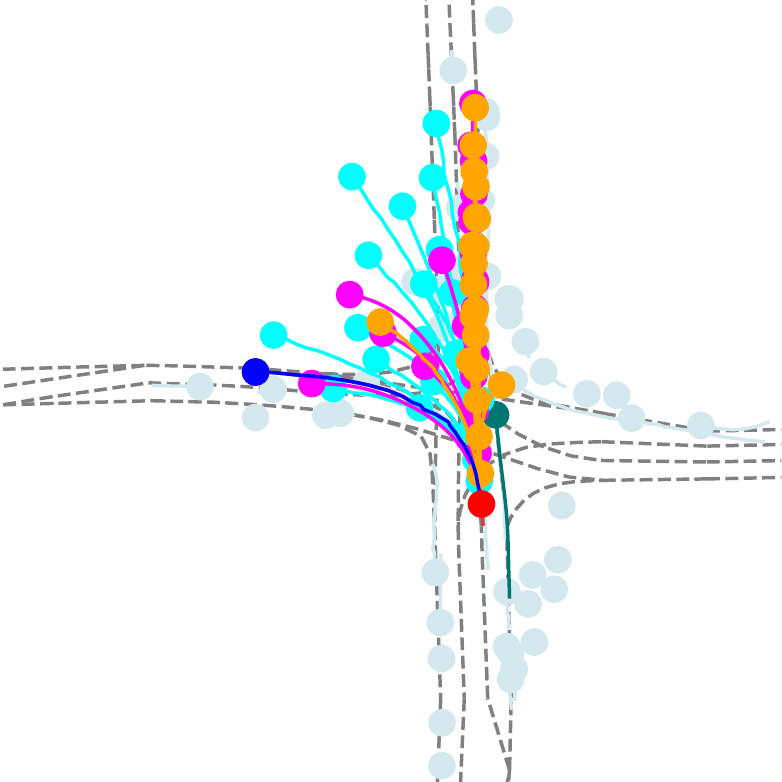} } 
\label{fig:vis_more_20}
\caption{More visual comparison results of 6 outputs (top row) and 20 outputs (bottom row) on HiVT-128}
\end{figure*}

\section{More Implement Details}
In this section, we provide more details about the implementation for each baseline.

For the feature distillation of VectorNet \cite{gao2020vectornet}, there is no clear agent branch and map branch. Therefore for VectorNet, we calculate the equivalent
map feature $f^e_m$ and the equivalent
agent feature $f^e_a$ in the global graph. The output prediction target agent feature $f^{global}$ of the global attention graph in VectorNet takes the subgraph output $f^{sub}$ as input, and can be written as follows:
\begin{equation}
\centering
\begin{aligned}
    f^{global} &= \sum (w_a^i * {f^{sub}_{a}}^i) + \sum( w_m^i * {f^{sub}_{m}}^i) \\
               &=\sum w_a^i* \sum (\frac{w_a^i}{\sum w_a^i} * {f^{sub}_{a}}^i)  + \sum w_m^i*\sum (\frac{w_m^i}{\sum{w_m^i}} * {f^{sub}_{m}}^i),\\
\end{aligned}
\label{eq：graph}
\end{equation}
where
\begin{equation}
\begin{aligned}
    \sum w_a^i + \sum w_m^i = 1.
\end{aligned}
\end{equation}

In equation \ref{eq：graph}, $f^{sub}_a$ means the agent trajectory feature and $f^{sub}_m$ means the map feature after the subgraph encoding. $w_a$ and $w_s$ mean the attention weights for agent features and map features respectively.
Equation \ref{eq：graph} can be seen as the weighted sum of an equivalent agent feature and an equivalent map feature. The equivalent
agent feature $f^e_a$  and the equivalent map feature $f^e_m$ can be written as follows, which can be seen as calculated by a separate agent encoding branch and a separate map encoding branch with $\frac{w_a}{\sum{w_a^i}}$ and $\frac{w_m}{\sum{w_m^i}}$ as the attention weight respectively:
\begin{equation}
\begin{aligned}
    f^e_a = \sum (\frac{w_a^i}{\sum{w_a^i}} * {f^{sub}_{a}}^i),\\
    f^e_m = \sum (\frac{w_m^i}{\sum{w_m^i}} * {f^{sub}_{m}}^i),\\
    \sum \frac{w_a^i}{\sum{w_a^i}} = 1,\\
    \sum \frac{w_m^i}{\sum{w_m^i}} = 1.\\
\end{aligned}
\end{equation}
We make the student agent feature ${f^e_a}^s$, the student map feature ${f^e_m}^s$, and the student fusion feature ${f^{global}}^s$ match the corresponding teacher feature ${f^e_a},{f^e_a},{f^{global}}$.
For the student, We use another agent attention branch to replace the map branch:
\begin{equation}
    {f^e_m}^s = \sum (\frac{a^{'i}_m}{\sum{a^{'i}_m}} * {{f^{sub}_{a}}^{s}}^i),
\end{equation}
where ${f^e_m}^s$ and ${f^{sub}_{a}}^s$ is the feature correcponds to ${f^e_m}$ and ${f^{sub}_{a}}$ in the student network. $a^{'}_m$ is the agent attention weight to the target agent.

For the output distillation of VectorNet, we take the binary cross entropy as in Y-Net \cite{mangalam2021goals} between rendered student goal probability $\overline{P^s}(\mathbf{g})$ and teacher goal probability ${P^t}(\mathbf{g})$ as described in the main paper.  
For a goal point $g$ in the selected goal set $\mathbf{g}$, the rendered probability of the regression-based student $\overline{P^s}$ is as follow:
\begin{equation}
\overline{P^s}(g) = P_{max}(g|\mu_{max},\sigma_{max})
\end{equation}
where the $P_{max}$ denotes the output hypothesis which has the maximum probability for $g$. And the teacher probability is renormalized as follow:
\begin{equation}
{P^t}(g) = \frac{P_{t}(g)}{\sum_{g^{'} \in \mathbf{g}} P_{t}(g^{'})}
\end{equation}
where $P_{t}$ is the original output probability of the teacher network. For VectorNet we use a single NVIDIA GeForce RTX 3090 GPU to train for 25 epochs with the Adam optimizer, the learning rate is initially set as 1e-3 and decay to 0.3 times every 5 epochs.

For the feature distillation of LaneGCN \cite{liang2020learning}, without the map we cannot get the lane graph connection information. So we just remove the student map graph convolution network and make the student agent features $f^s_a$ after the A-A interaction match the corresponding teacher feature $f^t_a$. For the output distillation of LaneGCN, we add a variance output branch to make the output Gaussian and take the Gaussian distillation as described in the main paper. For LaneGCN we use 2 NVIDIA GeForce RTX 3090 GPUs to train for 80 epochs with the Adam optimizer. The initial learning rate is set as 1e-3, and the decay to 1e-4 after 60 epochs.

For the feature distillation of HiVT \cite{zhou2022hivt}, we perform distillation on the agent feature $f_a$ after the agent temporal encoder in the local encoder, the map feature $f_m$ after the A-L interaction module and before the A-L update process, the fused agent feature after the A-L update process $f_{fusion}$, and the feature after the global encoder $f_{global}$. For the output distillation of HiVT, we take the Laplacian distillation according to the output distribution form. For HiVT we use 1 NVIDIA GeForce RTX 3090 GPU to train for 64 epochs and use the AdamW optimizer. The learning rate is initially set as 5e-4 and weight decay is set as 1e-4. The learning rate is decayed using the cosine annealing scheduler \cite{loshchilov2016sgdr} with $T=64$.

\section{Failure Case Study}
 In figure \ref{fig:fail} we show some failure cases of our proposed method. We can see from the figure when there are few agents on the future turning road, our FOKD might not recognize the road structure and then cannot make the turning prediction. Also, our method might be unable to capture the road semantics in some hard cases, therefore might fail to predict sudden acceleration.
 
\begin{figure*}[h]
\centering
\subfigure{ \includegraphics[width=0.3\linewidth]{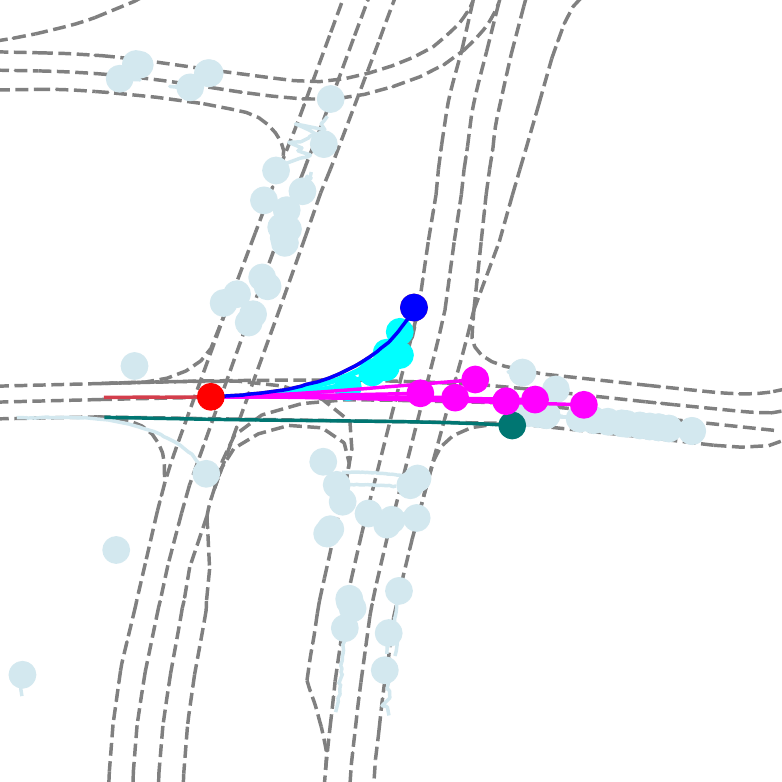}}
\subfigure{ \includegraphics[width=0.3\linewidth]{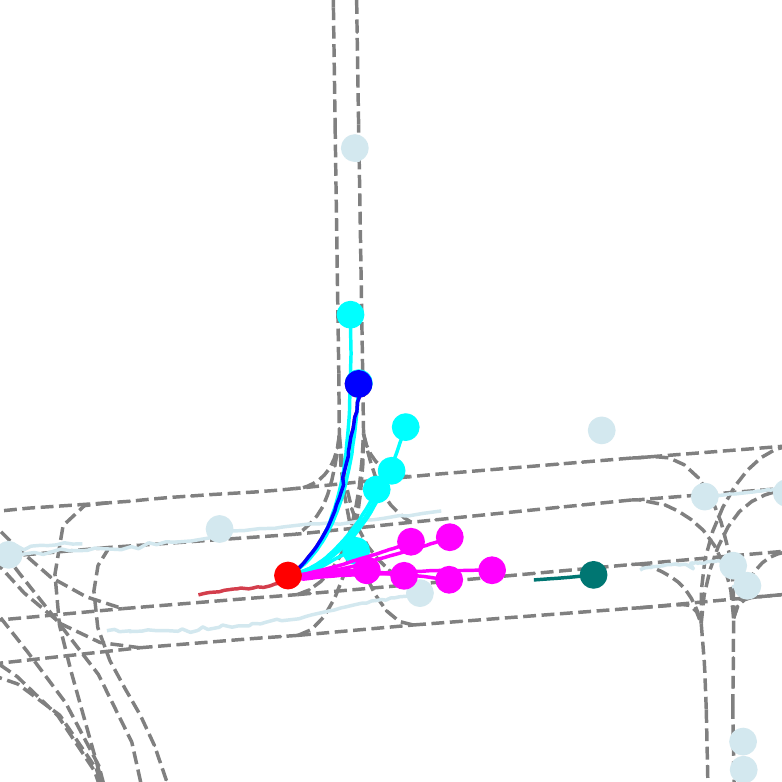}}
\subfigure{ \includegraphics[width=0.3\linewidth]{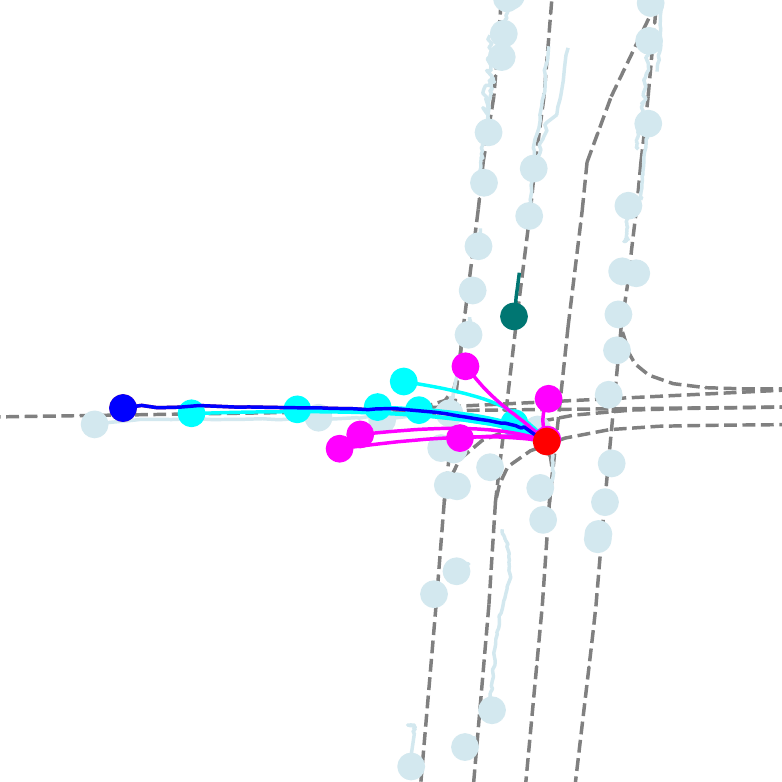}}
\caption{Visualization of some failure cases of FOKD on HiVT-128}
\label{fig:fail}
\end{figure*}

\section{Error Bars}
We implement our FOKD for 3 different random seeds on HiVT-128, and the averaged values of the metrics and the error bars are in table \ref{tab: error}.

\begin{table*}[h!]
\footnotesize
\resizebox{0.999\textwidth}{!}{
\centering
\begin{tabular}{lllllll}
\toprule
&$ADE_{6}$  & $FDE_{6}$  & $MR_{6}$ & $ADE_{20}$ & $FDE_{20}$ & $MR_{20}$  \\
\midrule
HiVT-128 + FOKD &0.71 $\pm$ 0.002& 1.11 $\pm$ 0.005 & 0.11 $\pm$ 0.004 & 0.61$\pm$ 0.002& 0.70$\pm$ 0.002& 0.05$\pm$ 0.002\\
\bottomrule
\end{tabular}
}
\caption{Averaged values of the metrics and error bars of our method with different random seeds.}
\label{tab: error}
\end{table*}

\section{Boarder Impact}
In this paper, we propose an original knowledge distillation method to enhance the performance of mapless trajectory prediction. Our approach differs from existing pipelines in that it takes teacher knowledge as supervision. Our method is useful in promoting prediction accuracy under challenging conditions, such as rough terrains and remote areas where HD maps are not available. As a result, it can help to improve the safety of autonomous driving.

However, several potential issues should be taken into consideration when our method is applied in real-world
scenarios. Firstly, similar to other learning-based prediction methods, there still remain concerns about interpretability
and robustness. Secondly, our knowledge distillation approach requires extra training epochs to pretrain a teacher network, which can be inefficient and lead to increased electricity consumption during the training phase. Finally, as with other learning-based prediction methods, training on manually collected datasets can result in biased network features that may adversely affect real-world performance during testing.

\end{document}